\documentclass[10pt,journal,compsoc]{IEEEtran}
\usepackage{graphicx}
\usepackage{bbding}
\usepackage{multirow}
\usepackage{algorithm}
\usepackage{algorithmic}
\usepackage{subfigure}
\usepackage{amsfonts}
\usepackage{mathtools, amssymb}
\usepackage{hyperref}
\usepackage{ragged2e}
\usepackage{color}

\newcommand{\tabincell}[2]{\begin{tabular}{@{}#1@{}}#2\end{tabular}}

\ifCLASSOPTIONcompsoc
  \usepackage[nocompress]{cite}
\else
  \usepackage{cite}
\fi

\ifCLASSINFOpdf
\else
\fi

\begin{document}
%
\title{Fast Class-wise Updating for Online Hashing}

\author{Mingbao Lin,
        Rongrong Ji*,~\IEEEmembership{Senior Member,~IEEE,}
        Xiaoshuai Sun,
        Baochang Zhang,
        Feiyue Huang, 
        Yonghong Tian, ~\IEEEmembership{Senior Member,~IEEE,}
        and~Dacheng Tao,~\IEEEmembership{Fellow,~IEEE}
\IEEEcompsocitemizethanks{\IEEEcompsocthanksitem M. Lin, R. Ji (Corresponding author) and X. Sun are with the Media Analytics and Computing Laboratory, Department of Artificial Intelligence, School of Informatics, Xiamen University, 361005, China.\protect\\
E-mail: rrji@xmu.edu.cn. \protect
%
%
\IEEEcompsocthanksitem R. Ji is also with Institute of Artificial Intelligence, Xiamen University, China.\protect
\IEEEcompsocthanksitem B. Zhang is with Beihang University, China.\protect
\IEEEcompsocthanksitem F. Huang is with Youtu Laboratory, Tencent, Shanghai, 200233, China.\protect
\IEEEcompsocthanksitem Y. Tian is with the Department of Computer Science and Technology, Peking University, Beijing, 100871, China.\protect
%
%
\IEEEcompsocthanksitem D. Tao is with the School of Computer Science, in the Faculty of Engineering, at The University of Sydney, 6 Cleveland St, Darlington, NSW 2008, Australia.\protect
%
}
\thanks{Manuscript received April 19, 2005; revised August 26, 2015.}}

\markboth{IEEE Transactions on Pattern Analysis and Machine Intelligence, vol. X, No. X, MMMMMMM YYYY}%
{Shell \MakeLowercase{\textit{et al.}}: Bare Demo of IEEEtran.cls for Computer Society Journals}

\IEEEtitleabstractindextext{%
\begin{abstract}
\justifying
Online image hashing has received increasing research attention recently, which processes large-scale data in a streaming fashion to update the hash functions on-the-fly.
To this end, most existing works exploit this problem under a supervised setting, \emph{i.e.}, using class labels to boost the hashing performance, which suffers from the defects in both adaptivity and efficiency:
First, large amounts of training batches are required to learn up-to-date hash functions, which leads to poor online adaptivity.
Second, the training is time-consuming, which contradicts with the core need of online learning.
In this paper, a novel supervised online hashing scheme, termed \textbf{F}ast \textbf{C}lass-wise Updating for \textbf{O}nline \textbf{H}ashing (FCOH), is proposed to address the above two challenges by introducing a novel and efficient inner product operation.
To achieve fast online adaptivity, a class-wise updating method is developed to decompose the binary code learning and alternatively renew the hash functions in a class-wise fashion, which well addresses the burden on large amounts of training batches. Quantitatively, such a decomposition further leads to at least 75\% storage saving.
To further achieve online efficiency, we propose a semi-relaxation optimization, which accelerates the online training by treating different binary constraints independently.
Without additional constraints and variables, the time complexity is significantly reduced. Such a scheme is also quantitatively shown to well preserve past information during updating hashing functions. We have quantitatively demonstrated that the collective effort of class-wise updating and semi-relaxation optimization provides a superior performance comparing to various state-of-the-art methods, which is verified through extensive experiments on three widely-used datasets.
\end{abstract}

}

\maketitle

\IEEEdisplaynontitleabstractindextext

%
\IEEEpeerreviewmaketitle

\IEEEraisesectionheading{\section{Introduction}\label{introduction}}
\IEEEPARstart{T}{raditional} hashing methods \cite{weiss2009spectral,liu2011hashing,gui2018r,gong2011iterative,wang2012semi,zhu2013linear,gong2013iterative,liu2015sequential,shen2015supervised,yu2016binary,long2017zero,zhang2017semi,liu2018dense,long2018transferable,song2018self,shen2018unsupervised,wang2018survey,cao2018deep,liu2018ordinal,he2018hashing,jin2018deep,wu2019deep,cakir2019hashing} are mostly designed to learn hash functions offline from a fixed collection of training data with/without supervised labels.
However, such a setting cannot handle dynamic application scenarios where data are fed into the system in a streaming fashion.
Therefore, online hashing has attracted much research attention recently \cite{huang2013online,lin2020hadamard,leng2015online,cakir2015adaptive,xie2016online
,lin2020similarity,cakir2017online,qi2017online,xie2017dynamic,fatih2017mihash,Chen2017FROSHFO,lin2018supervised,lin2019towards,yao2019online}.
It aims to online update the hash functions from the sequentially arriving data instances, which merits in its superior adaptivity and scalability for large-scale online retrieval applications.
Ideally, online hashing should efficiently update hash functions based on the streaming data on-the-fly, while preserving the information from the past data stream.
Existing works on online hashing can be classified into supervised and unsupervised methods.
Representative supervised methods include, but are not limited to, OKH \cite{huang2013online}, AdaptHash \cite{cakir2015adaptive}, OSH \cite{cakir2017online}, MIHash \cite{fatih2017mihash}, HCOH \cite{lin2018supervised} and BSODH \cite{lin2019towards}, which use supervised labels to guide the online hashing learning.
In contrast, unsupervised online hashing approaches, \emph{e.g.}, SketchHash \cite{leng2015online} and FROSH \cite{Chen2017FROSHFO} consider a smaller data sketch from a large dataset \cite{liberty2013simple} to update hash functions while preserving the main property of the dataset.
Due to the usage of label information, supervised online hashing methods usually yield better results over unsupervised ones, which therefore role as the main trend in the literature and are the focus of this paper.

\begin{figure}[!t]
\begin{center}
\includegraphics[height=0.53\linewidth]{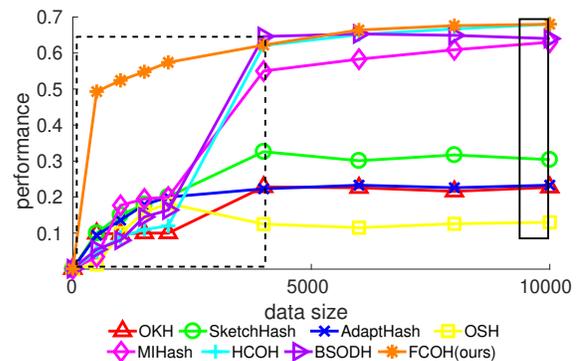}
\vspace{-1em}
\caption{\label{illustration}
Comparisons of the existing supervised online hashing methods \cite{huang2013online,leng2015online,cakir2015adaptive,cakir2017online,fatih2017mihash,Chen2017FROSHFO,lin2018supervised,lin2019towards} and ours in terms of adaptivity.
Existing methods reach the full performance only at the final training stage (solid rectangle box), which cannot be fast adapted online, and require a large number of training batches to learn up-to-date hash functions (dashed rectangle box).
Differently, the proposed FCOH method can achieve superior adaptivity online at the earlier stages with much less training data.
}
\end{center}
\vspace{-2em}
\end{figure}

%
However, supervised online hashing remains as an open problem due to two issues, as validated in Sec.\,\ref{experiment}:
First, a large number of training batches are required to learn up-to-date hash functions, which suffers poor online adaptivity.
Previous works \cite{huang2013online,leng2015online,cakir2015adaptive,cakir2017online,fatih2017mihash,Chen2017FROSHFO,lin2018supervised,lin2019towards} simply focus on designing hash models to obtain the full performance at the final training stage, which thereby lack sufficient adaptivity as shown in Fig.\,\ref{illustration}.
In comparison, such methods perform relatively poorly at the early training stages.
Second, the state-of-the-art supervised methods are mostly time-consuming to update each incoming data stream \cite{fatih2017mihash,lin2018supervised,lin2019towards}\footnote{This statement is quantitatively validated in Sec.\,\ref{training_efficiency}.}.
In particular, given a query, MIHash \cite{fatih2017mihash} has to calculate the Hamming distance between its neighbors and non-neighbors to update the hashing functions, which inevitably increases the burden on time consumption.
In \cite{lin2018supervised}, a pre-defined ECOC codebook \cite{horadam2012hadamard} is used to guide the learning of hash functions, and hence eliminates the strong constraints on similarity preserving.
However, the quality of short hashing codes still suffers poor performance due to the use of LSH to keep the consistency of code length and the size of ECOC codebook.
The work in \cite{lin2019towards} considers the inner product based scheme, which adopts two equilibrium factors to solve the ``data imbalance" problem and enables the usage of discrete optimization \cite{shen2015supervised} in an online setting.
However, on one hand, it introduces more variables to be optimized.
On the other hand, the usage of discrete optimization faces the convergence problem.
These two defects inevitably add more training burden.
OSH \cite{cakir2017online} adopts an Online Boosting algorithm \cite{babenko2009family} to enhance the online learning ability.
However, the usage of boosting inevitably causes training inefficiency.
It is worth to note that, the earlier supervised methods like OKH \cite{huang2013online} and AdaptHash \cite{cakir2015adaptive} and unsupervised methods like SketchHash  \cite{leng2015online} and FROSH \cite{Chen2017FROSHFO} are of high training efficiency, while their performance is far from satisfactory, which becomes worse when the training data increases (as quantitatively shown in Fig.\,\ref{illustration} and validated later in Sec.\,\ref{results}).

\begin{figure}[!t]
\begin{center}
\includegraphics[height=0.63\linewidth]{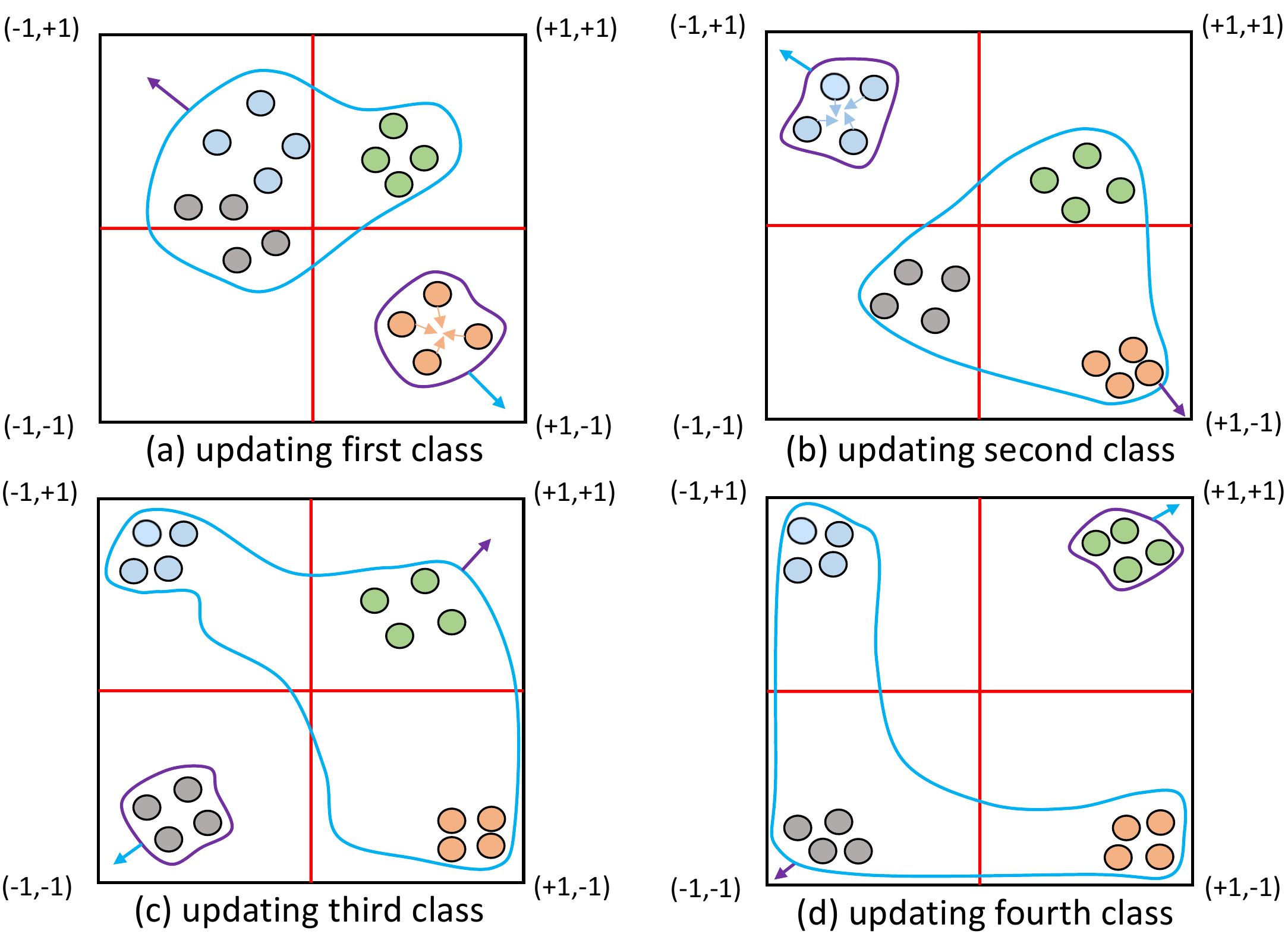}
\vspace{-1em}
\caption{\label{framework}
An illustration of the proposed class-wise updating.
For an arriving stream batch,
the updating is divided into several independent subprocesses, \emph{i.e.}, (a), (b), (c) and (d).
In each subprocess, FCOH aims to discriminate a particular class from the others,
which is conducted sequentially \big((a)$\rightarrow$(b)$\rightarrow$(c)$\rightarrow$(d)\big).
By this, the neighborhood information is well embedded in the Hamming space.
}
\end{center}
\vspace{-2em}
\end{figure}

%
To address the above issues, in this paper, we propose a novel supervised online hashing method, dubbed Fast Class-wise Updating for Online Hashing (FCOH).
Our key innovation is introducing the usage of inner product to preserve the similarity relationship in the Hamming space, which has been demonstrated to be very effective in learning binary codes offline \cite{liu2012supervised,shen2015learning,zhang2017semi,jiang2018asymmetric,lin2019towards}.
Unlike previous inner product based online hashing \cite{lin2019towards}, we address the defects in online adaptivity and training efficiency in a unified framework.
In this framework, we first develop a new updating scheme that alternatively renews the hash functions in a class-wise fashion, the advantages of which are illustrated in Fig.\,\ref{framework}. At each round, we only update samples from one class, which are well discriminated from the other classes.
Therefore, hash functions can be fast adapted online with much less training data.
Such a class-wise updating scheme differs from previous works that require all data involved in the training and therefore avoids large memory consumption.
Quantitatively, such an inner product based class-wise updating can dramatically reduce the space complexity by at least $75\%$.
Second, to accelerate the online training, we propose a semi-relaxation optimization which solves the binary constraints based on a divide-and-conquer method, \emph{i.e.}, relaxing the part of the binary constraint as a continuous constraint, while considering the rest as a constant that can be pre-computed by the hash weight learned in the previous stage.
In this way, no extra constraints and variables are involved, which highly reduces the training complexity.
Thus, the proposed FCOH is more efficient with higher scalability, and can be well applied to large-scale applications with lower complexity than prior works.
Besides, we show that the semi-relaxation optimization can well preserve the past information, which further boosts the performance.

The main contributions of our FCOH include:
\begin{itemize}
  \item We introduce the usage of inner product to preserve the similarity relationship in the Hamming space.
      Different from previous inner product based online methods, the key innovation of our framework lies in solving the online adaptivity and training efficiency problems in a unified framework.

  \item We develop a class-wise updating scheme to effectively update the hash functions with much less training data and storage consumption in online hashing.
      Unlike traditional online hash methods that reach full performance at the final training stage, our method can be fast adapted online, which well addresses the online adaptivity problem.

  \item We devise a new semi-relaxation optimization, which significantly reduces the training time of the inner product method.
      Our method solves the binary constraints based on a divide-and-conquer optimization, by relaxing part of the binary constraints to be a continuous constraint.
      The rest is formulated as a constant that can be pre-computed by the hash weights learned in the last stage, which well preserves the past information.

\end{itemize}

The collective effort of the proposed class-wise updating and semi-relaxation optimization greatly boosts the performance of online hashing.
Extensive experiments on three widely-used benchmarks, \emph{i.e.}, CIFAR-10, Places205 and MNIST, demonstrate that the proposed FCOH obtains superior accuracy and efficiency over the state-of-the-art methods \cite{huang2013online,leng2015online,cakir2015adaptive,cakir2017online,fatih2017mihash,Chen2017FROSHFO,lin2018supervised,lin2019towards}.

The rest of this paper is organized as follows:
In Sec.\,\ref{related_work}, we discuss the related work.
The proposed FCOH including class-wise updating and semi-relaxation optimization is elaborated in Sec.\,\ref{proposed_framework}.
Sec.\,\ref{experiment} reports the experimental results.
Finally, we conclude this work in Sec.\,\ref{conclusion}.

\section{Related Work}\label{related_work}

Supervised online hashing leverages label information to learn binary codes.
To our best knowledge, Online Kernel Hashing (OKH) \cite{huang2013online} is the first of this kind, which requires point pairs to update the hash functions via an online passive-aggressive strategy \cite{crammer2006online}.
Adaptive Hashing (AdaptHash) \cite{cakir2015adaptive} defines a hinge-like loss, which is approximated by a differentiable Sigmoid function to update the hash functions with SGD.
In \cite{cakir2017online}, a more general two-step hashing was introduced, in which binary Error Correcting Output Codes (ECOC) \cite{jiang2009efficient,kittler2001face,schapire1997using} are first assigned to labeled data, and then the hash functions are learned to fit the binary ECOC using online boosting.
Cakir \emph{et al}. \cite{fatih2017mihash} developed an Online Hashing with Mutual Information (MIHash), which targets at optimizing the mutual information between the neighbors and non-neighbors given a query.
Lin \emph{et al}. \cite{lin2018supervised} proposed a Hadamard Codebook based Online Hashing (HCOH), where a more discriminative Hadamard matrix \cite{horadam2012hadamard} is used as the ECOC codebook to guide the learning of hash functions.
Recently, Balanced Similarity for Online Discrete Hashing (BSODH) \cite{lin2019towards} was developed to investigate the correlation between new data and existing dataset via an inner product fashion.
To solve the ``data imbalance" problem, BSODH adopts two equilibrium factors to balance the similarity matrix and enables the application of discrete optimization \cite{shen2015supervised} in online learning.
Though extensive progress has been made in supervised online hashing \cite{huang2013online,cakir2015adaptive,cakir2017online,fatih2017mihash,lin2018supervised,lin2019towards}, there still remain two critical problems: poor online adaptivity and poor training inefficiency, as discussed in Sec.\,\ref{introduction}.

Unsupervised online hashing is mainly designed based upon the idea of ``data sketch" \cite{liberty2013simple}, where a small set of sketch data is used to preserve the main property of a large-scale dataset. %
To this end, Leng \emph{et al}. \cite{leng2015online} proposed an Online Sketching Hashing (SketchHash), which employs an efficient variant of SVD to learn hash functions, with a PCA-based batch learning on the sketch to learn hashing weights.
A faster version of Online Sketch Hashing (FROSH) was developed in \cite{Chen2017FROSHFO}, where the independent Subsampled Randomized Hadamard Transform (SRHT) is employed on different data chunks to make the sketch more compact and accurate, and further accelerate the learning process.
In general, unsupervised online hashing suffers low performance due to the lack of supervised labels.

\section{The Proposed Framework} \label{proposed_framework}

\subsection{Problem Definition} \label{definition}
Suppose the dataset is formed by a set of $n$ vectors, $\mathbf{X} = [\mathbf{x}_1, ..., \mathbf{x}_n] \in \mathbb{R}^{d \times n}$, accompanied by a set of class labels $ \mathbf{L} = [\mathbf{l}_1, ..., \mathbf{l}_n] \in \mathbb{N}^n$.
Besides, we denote $C = |set(\mathbf{L})|$ as the total categories of $\mathbf{L}$.
We aim to learn a set of $r$-bit hashing codes $\mathbf{B} = [\mathbf{b}_1, ..., \mathbf{b}_n] \in \{-1, +1\}^{r \times n}$ such that the desired neighborhood structure is preserved.
It is achieved by projecting the dataset $\mathbf{X}$ using a set of $r$ hash functions $H(\mathbf{X}) = \{h_i(\mathbf{X})\}_{i=1}^r$, \emph{i.e.},
\begin{equation}\label{hash_definition}
\mathbf{B} = H(\mathbf{X}) = sgn(\mathbf{W}^T\mathbf{X}),
\end{equation}
where $\mathbf{W} = \{ \mathbf{w}_i \}_{i=1}^r \in \mathbb{R}^{d \times r}$ is the projection matrix and $\mathbf{w}_i$ is the $i$-th hash function.
The sign function is defined as:
$$sgn(x)=
\begin{cases}
1& x > 1,\\
-1& \text{otherwise.}
\end{cases}$$

In the online setting, $\mathbf{X}$ comes in a streaming fashion.
Hence, for streaming data at the $t$-stage, we denote $\mathbf{X}^t = [\mathbf{x}_1^t, ..., \mathbf{x}_{n_t}^t] \in \mathbb{R}^{d \times n_t}$ as the input streaming data,
denote $\mathbf{B}^t = [\mathbf{b}_1^t, ..., \mathbf{b}_{n_t}^t] \in \{-1, +1\}^{r \times n_t}$ as the learned binary codes for $\mathbf{X}^t$,
and denote $\mathbf{L}^t = [\mathbf{l}_1^t, ..., \mathbf{l}_{n_t}^t] \in \mathbb{N}^{n_t}$ as the corresponding label set, where $n_t$ is the size of streaming data at the $t$-stage.
Further, we denote $C^t = |set(\mathbf{L}^t)|$ as the total categories received at the $t$-stage.
Correspondingly, the parameter $\mathbf{W}$ updated at the $t$-stage is denoted as $\mathbf{W}^t$.
Noticeably, $\mathbf{W}^t$ can only be deduced via $\mathbf{X}^t$ in an online setting.

In Tab.\,\ref{notation}, we summarize the notations used in the following contexts.

\begin{table}[]
\centering
\vspace{-1em}
\caption{List of Notations used in this paper.}
\label{notation}
\begin{tabular}{|c|c|}
\hline
Notation            & Meaning     \\
\hline
$\mathbf{X}$        &\tabincell{c}{$\mathbf{X}=[\mathbf{x}_1,...,\mathbf{x}_n]\in \mathbb{R}^{d \times n}$: set of $n$
                     training samples; \\ each column of $\mathbf{X}$ corresponds to one sample}   \\
\hline

$\mathbf{X}^t$       &\tabincell{c}{$\mathbf{X}=[\mathbf{x}^t_1,...,\mathbf{x}^t_{n_t}]\in \mathbb{R}^{d \times {n_t}}$: set of $n_t$ training samples at \\ the $t$-th stage; each column of $\mathbf{X}^t$ corresponds to one sample}   \\
\hline

$\mathbf{B}$        &\tabincell{c}{$\mathbf{B}=[\mathbf{b}_1,...,\mathbf{b}_n] \in \{-1, +1\}^{r \times n}$: binary code
                    matrix} for $\mathbf{X}$        \\
\hline

$\mathbf{B}^t$        &\tabincell{c}{$\mathbf{B}^t=[\mathbf{b}^t_1,...,\mathbf{b}^t_{n_t}]\in \{-1, +1\}^{r \times
                     {n_t}}$: binary code matrix} for $\mathbf{X}^t$        \\
\hline

$\mathbf{L}$        &\tabincell{c}{$\mathbf{L}=[\mathbf{l}_1,...,\mathbf{l}_n]\in\mathbb{N}^{n}$}: label set for
                    $\mathbf{X}$    \\
\hline

$\mathbf{L}^t$        &\tabincell{c}{$\mathbf{L}^t=[\mathbf{l}^t_1,...,\mathbf{l}^t_n]\in\mathbb{N}^{n_t}$:
                      label set for $\mathbf{X}^t$}    \\
\hline

$\mathbf{W}$         &\tabincell{c}{$\mathbf{W} = \{\mathbf{w}_i\}_{i=1}^r \in \mathbb{R}^{d \times r}$:
                     projection matrix \\ for the learned hashing functions}       \\
\hline

$\mathbf{W}^t$      &\tabincell{c}{$\mathbf{W}^t = \{\mathbf{w}^t_i\}_{i=1}^r \in \mathbb{R}^{d \times r}$:
                     projection matrix \\ updated at the $t$-th stage}       \\
\hline

$d$                 &feature dimension\\
\hline

$r$                 &number of the hashing bits       \\
\hline

$C$                   &total categories       \\
\hline

$C_t$                  &total categories at the $t$-th stage \\
\hline

$\| \cdot \|_F$            &Frobenius norm \\
\hline

$\| \cdot \|_1$            &$l_1$ norm \\
\hline

\end{tabular}
\vspace{-1.2em}
\end{table}

\subsection{The Proposed Method}\label{proposed_method}
The proposed framework is built based upon the inner product based formulation \cite{liu2012supervised,shen2015learning,zhang2017semi,jiang2018asymmetric,lin2019towards}.
The key ingredient is to map data points into binary codes, the inner product of which can well approximate the similarity matrix $\mathbf{S}^t \in \{-1, +1\}^{n_t \times n_t}$ where $\mathbf{S}^t_{ij} = 1$, if $\mathbf{l}_i^t = \mathbf{l}_j^t$, and $-1$ otherwise.
More precisely, the goal of inner product based methods is to minimize the following objective function:
\begin{equation} \label{inner_product}
   \mathcal{L} = \big\| (\mathbf{B}^t)^T\mathbf{B}^t - r\mathbf{S}^t \big\|_F^2 \quad s.t. \quad \mathbf{B}^t \in \{-1, +1\}^{r \times n_t},
\end{equation}
where $\|\cdot\|_F$ is the Frobenius norm.

The above formulation has been shown to be effective in traditional offline hashing methods \cite{liu2012supervised,shen2015learning,zhang2017semi,jiang2018asymmetric}.
However, directly applying this equation to online learning is infeasible due to the ``data imbalance" problem, as identified in \cite{lin2019towards}.
Specifically, at the $t$-th stage, Eq.\,(\ref{inner_product}) can be re-written as:

\begin{equation} \label{imbalance}
\begin{split}
&\mathcal{L}^t  = \underbrace{\sum_{i,j,\mathbf{S}_{ij}^t = 1}\big((\mathbf{b}_{i}^t)^T\mathbf{b}_{j}^t - r\big)^2}_{\text{term} \, \mathcal{A}} +
\underbrace{\sum_{i, j, \mathbf{S}_{ij}^t = -1}\big((\mathbf{b}_{i}^t)^T\mathbf{b}_{j}^t + r\big)^2}_{\text{term} \, \mathcal{B}}
\\& \qquad \qquad s.t. \quad {\mathbf{b}}^t_{i} \in \{-1, 1\}^r, {\mathbf{b}}^t_{j} \in \{-1,1\}^r.
\end{split}
\end{equation}

In an online setting, the similarity matrix $\mathbf{S}^t$ is highly sparse, \emph{i.e.}, most data pairs are dissimilar.
Therefore in practice, we have term $\mathcal{B}$ $>>$ term $\mathcal{A}$, which is referred to as the ``data imbalance'' problem.
Since the learning of binary codes heavily relies on such dissimilar pairs, the retrieval performance cannot be satisfied as expected.
To address this problem, the work in \cite{lin2019towards} introduces two equilibrium factors and enables the usage of discrete optimization in the online setting.
However, such a solution needs a large number of training batches to learn up-to-date hash functions, which causes poor online adaptivity at the early training stages, as illustrated in Fig.\,\ref{illustration}.
Besides, it also introduces more variables to be optimized, and the discrete optimization further brings about the convergence problem.
These two drawbacks unavoidably lead to the training inefficiency.
The novelty of our method lies in two-fold:
First, to solve the problem of poor online adaptivity, we propose a ``Class-wise Updating" scheme, which decomposes the optimization into several independent subprocesses with each responsible for one category, resulting in high accuracy and at least $75\%$ storage saving as shown latter in Sec.\,\ref{class_wise_updating}.
Second, to address the problem of training inefficiency, we propose a ``Semi-relaxation Optimization" scheme. In this scheme, one part of the binary constraints in Eq.\,(\ref{inner_product}) is relaxed as a continuous variable while the other is considered as a constant variable, through which the time complexity is drastically reduced.
We detailedly describe the above two methods as below.

\subsubsection{\textbf{Class-wise Updating}}\label{class_wise_updating}

\begin{figure}[!t]
\begin{center}
\includegraphics[height=0.5\linewidth]{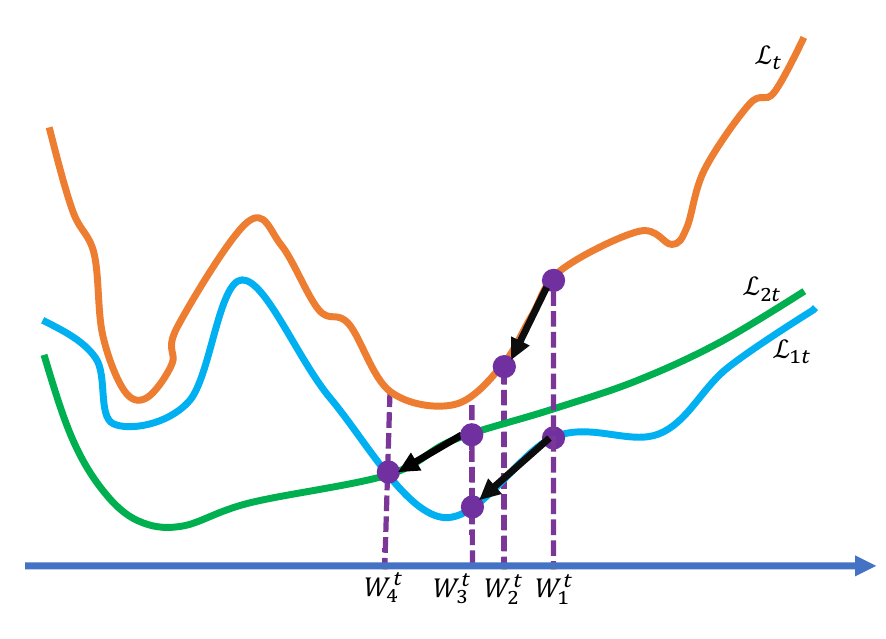}
\vspace{-1em}
\caption{\label{class_wise_gradient}
Analysis on the class-wise updating.
The blue and green curves represent the loss for the first class and the second class.
The orange curve stands for the overall loss.
$\mathbf{W}^t_1$ is the start point for the $t$-th updating stage.
Without class-wise updating, $\mathbf{W}_1^t$ is updated for only once and the renewed weights are $\mathbf{W}^t_2$, 
which is not a good choice for both the first class and second class.
With class-wise updating, the hash weights are updated gradient-by-gradient.
It first updates hash functions using $\mathcal{L}_{1t}$ and obtains $\mathbf{W}_3^t$, based on which $\mathcal{L}_{2t}$ is used and then $\mathbf{W}_{4}^t$ is obtained, serving as the final weights for the $t$-stage and shows an overall better result for class 1 and class 2.
}
\end{center}
\vspace{-2em}
\end{figure}

To solve the poor online adaptivity, we develop a class-wise updating scheme as shown in Fig.\,\ref{framework}.
Detailedly, we re-write Eq.\,(\ref{imbalance}) as follows:

\begin{equation}\label{class_wise}
\begin{split}
&\mathcal{L}^t = \\&
\sum_{c}^{C_t}\Big(  \underbrace{\sum_{i,j, \mathbf{l}^t_i = \mathbf{l}^t_j = c}\bigl((\mathbf{b}_i^t)^T\mathbf{b}_j^t - r\bigl)^2}_{term\,1}  + \underbrace{\sum_{i,j, \mathbf{l}^t_i = c, \mathbf{l}^t_j \neq c}\bigl((\mathbf{b}_i^t)^T\mathbf{b}_j^t + r\bigl)^2}_{term\,2}  \Big) \\&
\qquad \qquad \qquad s.t. \quad {\mathbf{b}}^t_{i} \in \{-1, 1\}^r, {\mathbf{b}}^t_{j} \in \{-1,1\}^r,
\end{split}
\end{equation}
where $c$ denotes the $c$-th class at the $t$-th training stage.
$Term\,1$ stands for the inner product among instances from the $c$-th class, and $term\,2$ represents the inner product between instances from the $c$-th class and instances from classes excluding the $c$-th class.
The basic idea behind class-wise updating is that at the $t$-th training stage, we decompose the training procedure into $C_t$ independent subprocesses as illustrated in Fig.\,\ref{framework}.
At the $c$-th subprocess, FCOH focuses on learning discriminative binary codes for the $c$-th class, which can well push the training instances of the $c$-th class away from instances from the other classes.
Fig.\,\ref{class_wise_gradient} illustrates a toy example that the class-wise updating renews the hash weights via a gradient-by-gradient fashion.
After a total of $C_t$ subprocesses, the final hash weights keep an overall better performance for all classes and obsesses a fast adaptivity to the coming data at the $t$-stage.
%
%
%

%
Without loss of generalization, we analyze that the decomposed Eq.\,(\ref{class_wise}) in the meantime can drastically reduce at least $75\%$ space complexity in Eq.\,(\ref{inner_product}).
To that effect, we first reformulate Eq.\,(\ref{class_wise}) in the following:
\begin{equation}\label{sum_form}
\mathcal{L}_t = \sum_{c}^{C_t}\mathcal{L}_{ct},
\end{equation}
where $\mathcal{L}_{ct} = term\,1 + term\,2$ in Eq.\,(\ref{class_wise}), and it represents the loss objective for the $c$-th class.
Further, we rewrite $\mathcal{L}_{ct}$ into a matrix form:
\begin{equation} \label{matrix_form}
\begin{split}
&\mathcal{L}_{ct} = \underbrace{\| (\mathbf{B}^{ct})^T\mathbf{B}^{ct} - r \|_F^2}_{term\,1}  +  \underbrace{\| (\mathbf{B}^{ct})^T\mathbf{B}^{\bar{c}t} + r \|_F^2}_{term\,2}
\\& s.t. \quad \mathbf{B}^{ct} \in \{-1, +1\}^{r \times n_{ct}}, \mathbf{B}^{\bar{c}t} \in \{-1, +1\}^{r \times n_{\bar{c}t}},
\end{split}
\end{equation}
where $\mathbf{B}^{ct}$ is the matrix consisting of binary codes from the $c$-th class and $\mathbf{B}^{\bar{c}t}$ is the matrix consisting of binary codes excluding the $c$-th class;
$n_{ct}$ is the total number of training instances from the $c$-th class at the $t$-stage and
$n_{\bar{c}t}$ is the total number of training instances excluding the $c$-th class at the $t$-stage.
It is easy to derive that the space complexity in Eq.\,(\ref{matrix_form}) is $\mathcal{O}(n_{ct}n_{\bar{c}t})$ instead of $\mathcal{O}(n_t^2)$ in Eq.\,(\ref{inner_product}).
The storage saving rate $R$ can be formulated as:
\begin{equation}\label{saving_rate}
\begin{split}
  &R = 1 - \frac{n_{ct}n_{\bar{c}t}}{n_t^2}
  \\  s.t&. \quad n_{ct} + n_{\bar{c}t} = n_t.
\end{split}
\end{equation}

When $n_{ct} = n_{\bar{c}t} = \frac{1}{2}n_t$, $n_{ct}n_{\bar{c}t} = \frac{1}{4}n_t^2$ requires the largest memory consumption.
In this case, the proposed FCOH obtains $R = 75\%$ storage saving rate.
When $n_{ct} \neq n_{\bar{c}t}$, $n_{ct}n_{\bar{c}t} < \frac{1}{4}n_t^2$ and $R > 75\%$.
Therefore, the proposed FCOH can reduce at least $75\%$ storage consumption.

We note that the early work \cite{rastegari2012attribute} adopts a similar class-wise updating scheme.
However, our work still has distinct difference to \cite{rastegari2012attribute}: 
First, \cite{rastegari2012attribute} solves the offline hashing problem while FCOH solves the online hashing problem.
Second, \cite{rastegari2012attribute} solves the offline hashing problem based on classifiers while FCOH solves the online hashing problem based the on inner product.
Lastly, \cite{rastegari2012attribute} proposes to class-wise innovate the binary codes, while FCOH proposes to class-wise update the hash weights.

%
%
%
%
%
%

\subsubsection{\textbf{Semi-relaxation Optimization}} \label{semi_relaxation_optimization}
Due to the discrete constraint in Eq.\,(\ref{matrix_form}), it is easy to derive that the above problem is highly non-convex and difficult (usually NP-hard) to solve.
An alternative and suboptimal solution is to apply the discrete cyclic coordinate descent (DCC) developed in \cite{shen2015supervised}, \emph{i.e.}, updating one hashing bit while fixing the other bits.
However, as shown in the recent online hashing method \cite{lin2019towards}, the discrete optimization brings more variables to be optimized
and the bit-wise scheme is hard to converge.
The above two issues inevitably lead to more training burden (as verified in Tab.\,\ref{training_time}), which should be further addressed in online hashing.

To achieve the efficient optimization, we adopt the relaxation method by partly removing the $sgn(x)$ function in Eq.\,(\ref{hash_definition}).
However, different from traditional relaxation process where the sign functions are all removed from the binary constraints, we propose to semi-relax the sign function in Eq.\,(\ref{matrix_form}) based on a divide-and-conquer idea.
It removes the sign function in part of the inner product of binary matrix, and regards the other part of binary matrix as a constant.

To that effect, we reformulate Eq.\,(\ref{matrix_form}) as follows:
\begin{equation} \label{semi_relax}
\begin{split}
&\hat{\mathcal{L}}_{ct} = {\lambda}_1\big\| \big((\mathbf{W}^{t-1})^T\mathbf{X}^{ct}\big)^T\mathbf{B}^{c(t-1)} - r \big\|_F^2
\\& \qquad + {\lambda}_2\big\| \big((\mathbf{W}^{t-1})^T\mathbf{X}^{ct}\big)^T\mathbf{B}^{\bar{c}(t-1)} + r \big\|_F^2,
\end{split}
\end{equation}
where
\begin{equation} \label{b_current}
\mathbf{B}^{c(t-1)} = sgn\big({(\mathbf{W}^{t-1}})^T\mathbf{X}^{ct}\big) \in \{-1, +1\}^{r \times n_{ct}},
\end{equation}
and
\begin{equation}\label{b_past}
\mathbf{B}^{\bar{c}(t-1)} = sgn\big({(\mathbf{W}^{t-1}})^T\mathbf{X}^{\bar{c}t}\big) \in \{-1, +1\}^{r \times n_{\bar{c}t}},
\end{equation}
which are \emph{two constant matrices} pre-computed at the $(t-1)$-th stage.
And ${\lambda}_1$ and ${\lambda}_2$ are two hyper-parameters to balance the importance of the two terms.
The advantages of such operations are in two-fold:
Firstly, $\mathbf{B}^{c(t-1)}$ and $\mathbf{B}^{\bar{c}(t-1)}$ are pre-computed by the $(t-1)$-th hash weights.
Therefore, not only training time is saved at the $t$-stage, but also more information is learned from the past stage.
Meanwhile, in updating, the term $\mathbf{W}^{t-1}\mathbf{X}^{ct}$ aims to provide new information from the current stage.
Therefore, the semi-relaxation can learn information from newly streaming data, and it can also preserve information from the last stage.
Secondly, compared with the inner product based method in \cite{lin2019towards}, such a semi-relaxation simplifies the optimization complexity, \emph{i.e.}, only $\mathbf{W}^t$ needs to be optimized, which greatly improves the training efficiency.

Nevertheless, Eq.\,(\ref{semi_relax}) only considers the clues from the current and the last stage.
As the training proceeds, the streaming data from early stages easily suffers large quantization errors.
However, quantizing all the past streaming data is time-consuming as the dynamic dataset grows.
Besides, it also violates the principle of online hashing since hashing functions should be updated only based on the newly arriving streaming data online.
To solve issue, inspired by\,\cite{liu2016deep}, we further propose to quantize the centre of the $c$-th class that can be readily tracked, as follows:
\begin{equation}\label{quantization}
  \big\| |\big((\mathbf{W}^{t-1})^T\mathbf{\bar{x}}^{ct}\big)| - \mathbf{1} \big\|_1,
\end{equation}
where $\|\cdot\|_1$ is the $\ell_1$-norm and $|\cdot|$ returns the absolute value, $\mathbf{1}$ is an all-one matrix and
\begin{equation}\label{center}
\mathbf{\bar{x}}^{ct} = \frac{N_{c(t-1)}\mathbf{\bar{x}}^{c(t-1)} + \sum_{i=1}^{n_{ct}}\mathbf{x}_i^{ct}}{N_{ct}},
\end{equation}
where $N_{ct} = N_{c(t-1)} + n_{ct}$ and $N_{c1} = n_{c1}$.
$N_{ct}$ denotes the number of the $c$-th class and $\mathbf{\bar{x}}^{ct}$ is the center of the $c$-th class.

Combining Eq.\,(\ref{semi_relax}) and Eq.\,(\ref{quantization}) leads the final objective function as below:
\begin{equation} \label{final}
\begin{split}
\tilde{\mathcal{L}}_{ct} &=\big\| |\big((\mathbf{W}^{t-1})^T\mathbf{\bar{x}}^{ct}\big)| - \mathbf{1} \big\|_1  \\&
+ {\lambda}_1\big\| \big((\mathbf{W}^{t-1})^T\mathbf{X}^{ct}\big)^T\mathbf{B}^{c(t-1)} - r \big\|_F^2 \\&
+  {\lambda}_2\big\| \big((\mathbf{W}^{t-1})^T\mathbf{X}^{ct}\big)^T\mathbf{B}^{\bar{c}(t-1)} + r \big\|_F^2,
\end{split}
\end{equation}

\begin{algorithm}[t]
\caption{Fast Class-wise Updating for Online Hashing (FCOH) \label{alg1}}
\renewcommand{\algorithmicrequire}{\textbf{Input:}}
\renewcommand{\algorithmicensure}{\textbf{Output:}}
\begin{algorithmic}[1]
\REQUIRE
Dataset $\mathbf{X}$ and its label $\mathbf{L}$, number of hash bits $r$, parameter $\mu$, ${\lambda}_1$ and ${\lambda}_2$.
\ENSURE
Hash codes $\mathbf{B}$ for $\mathbf{X}$ and projection coefficient matrix $\mathbf{W}$.\\
\STATE
Initialize $\mathbf{W}^0$ with the normal Gaussian distribution.
\FOR {$t =1 \to T$}
    \FOR {$c = 1 \to C^t$}
        \IF{$t = 1$}
            \STATE $N_{ct} = n_{ct}$;
            \STATE ${\bar{x}}^{ct} = \frac{\sum_{i=1}^{n_{ct}}\mathbf{x}_i^{ct}}{N_{ct}}$;
        \ELSE
            \STATE $N_{ct} = N_{c(t-1)} + n_{ct}$;
            \STATE Compute $\mathbf{\bar{x}}^{ct}$ via Eq.\,(\ref{center});
        \ENDIF
        \STATE Compute $\mathbf{B}^{c(t-1)}$ via Eq.\,(\ref{b_current});
        \STATE Compute $\mathbf{B}^{\bar{c}(t-1)}$ via Eq.\,(\ref{b_past});
        \STATE Compute $\frac{\partial \tilde{\mathcal{L}}_{ct}(\mathbf{W}^{t-1})}{\partial \mathbf{W}^{t-1}}$ via Eq.\,(\ref{derivative});
        \STATE Update $\mathbf{W}^t$ via Eq.\,(\ref{optimization});
    \ENDFOR
\ENDFOR
\STATE Set $\mathbf{W} = \mathbf{W}^t$;
\STATE Compute $\mathbf{B} = sgn(\mathbf{W}^T\mathbf{X})$;
\STATE Return $\mathbf{W}$ and $\mathbf{B}$.
\end{algorithmic}
\end{algorithm}
To optimize Eq.\,(\ref{final}), we adopt the SGD optimization to update the hash functions for the $c$-th class as follows:
\begin{equation}\label{optimization}
\mathbf{W}^t \leftarrow \mathbf{W}^{t-1} - {\mu}\frac{\partial \tilde{\mathcal{L}}_{ct}(\mathbf{W}^{t-1})}{\partial \mathbf{W}^{t-1}},
\end{equation}
where $\mu$ is the learning rate.
The derivative of $\tilde{\mathcal{L}}_{ct}(\mathbf{W}^{t-1})$ \emph{w.r.t}. $\mathbf{W}^{t-1}$ can be obtained as:
\begin{equation}\label{derivative}
\begin{split}
&\frac{\partial \tilde{\mathcal{L}}_{ct}(\mathbf{W}^{t-1})}{\partial \mathbf{W}^{t-1}} =
\bar{\mathbf{x}}^{ct}\sigma\big((\mathbf{W}^{t-1})^T{\bar{\mathbf{x}}^{ct}}\big)^T
\\&
+2{\lambda}_1\bigg(\mathbf{X}^{ct}\Big(\big((\mathbf{W}^{t-1})^T\mathbf{X}^{ct}\big)^T\mathbf{B}^{c(t-1)} - r\Big) \bigg)\big(\mathbf{B}^{c(t-1)}\big)^T
\\&
+2{\lambda}_2\bigg(\mathbf{X}^{ct}\Big(\big((\mathbf{W}^{t-1})^T\mathbf{X}^{ct}\big)^T\mathbf{B}^{\bar{c}(t-1)} + r\Big) \bigg)\big(\mathbf{B}^{\bar{c}(t-1)}\big)^T,
\end{split}
\end{equation}
where the $\sigma(x)$ function is defined as:
$$\sigma(x)=
\begin{cases}
1& x > 1 \quad or \quad -1 < x < 0,\\
-1& \text{otherwise.}
\end{cases}$$

Without loss of generality, the main procedures of the proposed FCOH are outlined in Alg.\,\ref{alg1}.

\subsection{\textbf{Time Complexity}} \label{time_complexity}
In Alg. \ref{alg1}, the majority of the training time is spent on the operations between line $11$ to line $13$ to update the hash functions on the $c$-th class at the $t$-stage.
Specifically, the complexity of calculating $\mathbf{B}^{c(t-1)}$ in line $11$ is $\mathcal{O}(rdn_{ct})$ and it is $\mathcal{O}(rdn_{\bar{c}t})$ for calculating $\mathbf{B}^{\bar{c}(t-1)}$ in line $12$.
The time cost for calculating derivative of $\tilde{\mathcal{L}}_{ct}(\mathbf{W}^{t-1})$ \emph{w.r.t}. $\mathbf{W}^{t-1}$ in line $13$ is $\mathcal{O}(rdn_{ct} + rn_{ct}^2 + dn_{ct}^2 + rn_{ct}n_{\bar{c}t} + dn_{ct}n_{\bar{c}t} + rdn_{\bar{c}t} + rd)$.
Since $r << d$ and $n_{ct} << n_{\bar{c}t}$, the total complexity is $\mathcal{O}(rdn_{\bar{c}t} + dn_{ct}n_{\bar{c}t})$.
We denote $e = \max(r, n_{ct})$ and the complexity can be re-written as $\mathcal{O}(edn_{\bar{c}t})$.
Hence, the overall time complexity at the $t$-th training stage is $\mathcal{O}(C_tedn_{\bar{c}t})$.
As can be seen in Sec.\,\ref{experiment}, $C_t$, $e$ and $n_{\bar{c}t}$ are small values.
Therefore, the time complexity of the proposed method mainly depends on the feature dimension $d$, which guarantees the efficiency and scalability of the proposed method.

\section{Experiments} \label{experiment}
To validate the model accuracy and learning efficiency, we compare our FCOH with several state-of-the-art methods
\cite{huang2013online,leng2015online,cakir2015adaptive,cakir2017online,fatih2017mihash,lin2018supervised,lin2019towards} on three widely-used datasets, \emph{i.e.}, CIFAR-10\cite{krizhevsky2009learning}, Places205 \cite{zhou2014learning}, and MNIST \cite{lecun1998gradient}.

\subsection{Experimental Settings} \label{experimental_setting}
\subsubsection{\textbf{Datasets}} \label{datasets}
\emph{CIFAR-10} consists of $60,000$ images from $10$ classes.
Each class contains $6,000$ instances and each image is represented as a $4,096$-dim CNN vector \cite{simonyan2014very}.
As in \cite{fatih2017mihash,lin2018supervised,lin2019towards}, we randomly select $59,000$ samples to form a retrieval set. And the remaining $1,000$ are used to construct a test set.
Besides, we randomly sample $20K$ images from the retrieval set as the training set to learn hash functions in a streaming fashion.

\emph{Places205} is a large-scale dataset \cite{zhou2014learning} that contains $2.5$ million images from $205$ scene categories.
The feature of each image is extracted from the AlexNet \cite{krizhevsky2012imagenet} and then reduced to a $128$-dim feature by PCA.
Following \cite{lin2018supervised}, $20$ instances from each category are randomly sampled to construct the test set, and the others are used to form the retrieval set.
Lastly, a random subset of $100K$ images from the retrieval set is used to learn the hash functions in a streaming fashion.

\emph{MNIST} contains $70,000$ handwritten digit images from $0$ to $9$ \cite{lecun1998gradient}.
Each image is represented by $28 \times 28 = 784$-dim normalized original pixels.
According to the experimental setting in \cite{lin2019towards}, we construct the test set by randomly sampling $100$ instances from each class.
The rest is used to form the retrieval set.
Finally, a random subset of $20,000$ images from the retrieval set is randomly sampled to form the training set to learn the hash functions in a streaming fashion.

The detailed analysis on the size of data stream, \emph{i.e.}, $n_t$, is given in Sec.\,\ref{n_t}.

\begin{table}[]
\centering
\caption{Parameter Configurations on Three Benchmarks. \label{configuration}}
\vspace{-1em}
\begin{tabular}{c|c|c|c}
\hline
Method     & CIFAR-10              &Places205        &MNIST \\
\hline
$n_t$           &$100$                 &$1,000$          &$100$ \\
\hline
${\lambda}_1$   &1e-2                   &1e-1             &1e-1      \\
\hline
${\lambda}_2$   &1e-2                   &1e-1             &1e-2   \\
\hline
$\mu$           &1e-1                   &1e-4             &1e-2      \\
\hline

\end{tabular}
\vspace{-1em}
\end{table}

\subsubsection{\textbf{Evaluation}} \label{evaluation}
Following the previous methods \cite{lin2018supervised,lin2019towards}, we adopt the following protocols to evaluate the performance:
mean Average Precision (denoted as \textit{mAP}),
Precision within a Hamming ball of radius $2$ centered on each query (denoted as Precision@H$2$),
\emph{m}AP vs. different sizes of training instances curves and their corresponding areas under the \emph{m}AP curves (denoted as AUC),
Precision of the top K retrieved neighbors curves (denoted as Precision@K) and their corresponding areas under the Precision@K curves (denoted as AUC).
For Precision@K, the values of K are ranged in $\{1, 5, 10, 20, 30, ..., 90, 100 \}$ on all benchmarks, since the values of $K \le 100$ are common choices and are closer to the user behavior when investigating the search results.
Regarding \emph{m}AP \emph{vs}. different sizes of training instances, on CIFAR-10 and MNIST, the experimental results are sampled when the size of training instances falls in $\{2K, 4K, 6K, ..., 18K, 20K\}$, and it is $\{5K, 10K, ..., 95K, 100K\}$ on Places205 due to its large scale.
Such settings provide a fair comparison with state-of-the-art method, \emph{i.e.}, BSODH \cite{lin2019towards}, where the best choice of streaming data size is $2K$, $2K$ and $5K$ on CIFAR-10, MNIST and Places205, respectively.

Notably, when reporting the \emph{m}AP performance on Places205, following the works in \cite{fatih2017mihash,lin2018supervised,lin2019towards}, we only compute the top $1,000$ retrieved items (denoted as \emph{m}AP@1,000) due to its large scale and heavy time consumption.
The above metrics are evaluated with hashing bits varying among $8$, $16$, $32$, $48$, $64$ and $128$.

\subsubsection{\textbf{Baselines}} \label{baselines}
To illustrate the effectiveness of the proposed FCOH, we compare our method with several state-of-the-art online hashing algorithms, including
Online Kernel Hashing (OKH) \cite{huang2013online},
Online Sketching Hashing (SketchHash) \cite{leng2015online},
Adaptive Hashing (AdaptHash) \cite{cakir2015adaptive},
Online Supervised Hashing (OSH) \cite{cakir2017online},
Online Hashing with Mutual Information (MIHash) \cite{fatih2017mihash},
Hadamard Codebook based Online Hashing (HCOH) and
Balanced Similarity for Online Discrete Hashing (BSODH) \cite{lin2019towards}.
The source codes of these methods are publicly available.
Our model is implemented with MATLAB.
The experiments are conducted on a server with a $3.60$GHz Intel Core I$7$ $4790$ CPU and $16$G RAM.

\begin{table*}[!t]
\centering
\caption{\emph{m}AP And Precision@H$2$ Comparisons on CIFAR-10 with 8, 16, 32, 48, 64 And 128 Bits. %
The Best Result Is Labeled with Boldface And The Second Best Is with An Underline.}
\vspace{-1em}
\label{map_precision_cifar}
\begin{tabular}{c|cccccc|cccccc}
\hline
\multirow{2}{*}{Method} & \multicolumn{6}{c|}{\emph{m}AP}                   & \multicolumn{6}{c}{Precision@H$2$}          \\
\cline{2-13}
                        & 8-bit & 16-bit & 32-bit &48-bit & 64-bit & 128-bit & 8-bit & 16-bit & 32-bit &48-bit & 64-bit
                        &128-bit \\
\hline
OKH                     &0.100  &0.134   &0.223   &0.252  &0.268  &0.350    &0.100  &0.175   &0.100   &0.452  &0.175
                        &0.372  \\
\hline
SketchHash              &0.248  &0.301   &0.302   &0.327  &  -    &   -     &0.256  &0.431   &0.385   &0.059  &   -
                        & -     \\
\hline
AdaptHash               &0.116  &0.138   &0.216   &0.297  &0.305  &0.293    &0.114  &0.254   &0.185   &0.093  &0.166
                        &0.164      \\
\hline
OSH                     &0.123  &0.126   &0.129   &0.131  &0.127  &0.125    &0.120  &0.123   &0.137   &0.117  &0.083
                        &0.038 \\
\hline
MIHash                  &0.512  &0.640   &0.675   &0.668  &0.667  &0.664    &0.170  &0.673   &0.657   &0.604  &0.500
                        &0.413   \\
\hline
HCOH                    &0.536  &\textbf{0.698}   &0.688 &\underline{0.707} &\underline{0.724}  &\underline{0.734}
                        &\underline{0.333}  &\underline{0.723}   &\underline{0.731} &0.694   &0.633  &0.471 \\
\hline
BSODH                   &\underline{0.564}  &0.604   &\underline{0.689} &0.656  &0.709  &0.711    &0.305  &0.582   &0.691 &\underline{0.697}  &\textbf{0.690}
                        &\underline{0.602}  \\
\hline
\hline
FCOH                    &\textbf{0.602}&\underline{0.674}&\textbf{0.702}&\textbf{0.711}&\textbf{0.739}&\textbf{0.742}
                        &\textbf{0.484}&\textbf{0.738}&\textbf{0.743}&\textbf{0.711}&\underline{0.648}&\textbf{0.618}\\
\hline
\end{tabular}
\vspace{-1.5em}
\end{table*}

\begin{figure}[!t]
\begin{center}
\includegraphics[height=0.57\linewidth]{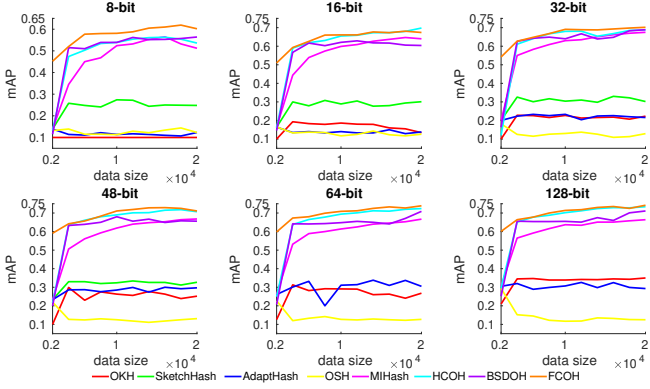}
\vspace{-2em}
\caption{\label{map_instance_cifar} \emph{m}AP performance with respect to different sizes of training instances on CIFAR-10.}
\end{center}
\vspace{-0.5em}
\end{figure}

%
\subsubsection{\textbf{Settings}}\label{settings}
To share the same dataset configurations on all the three benchmarks, we directly adopt the parameters as described in \cite{lin2018supervised,lin2019towards}, which have been carefully validated for each method.
The following elaborates the parametric settings.
\begin{itemize}
\item \textbf{OKH}: The tuple ($C, \alpha$) is set as (0.001, 0.3), (0.0001, 0.7) and (0.001,0.3) on CIFAR-10, Places205 and MNIST, respectively.
\item \textbf{SketchHash}: The tuple ($sketch size, batch size$) is set as (200, 50), (100, 50) and (200, 50) on CIFAR-10, Places205 and MNIST, respectively.
\item \textbf{AdaptHash}: The tuple ($\alpha, \lambda, \eta$) is set as (0.9, 0.01, 0.1), (0.9, 0.01,0.1) and (0.8, 0.01, 0.2) on CIFAR-10, Places205 and MNIST, respectively.
\item \textbf{OSH}: On all datasets, $\eta$ is set as 0.1 and the ECOC codebook $C$ is populated the same way as in \cite{cakir2017online}.
\item \textbf{MIHash}: The tuple ($\theta, \mathcal{R}, A$) is set as (0, 1000, 10), (0, 5000, 10) and (0, 1000, 10) on CIFAR-10, Places205 and MNIST, respectively.
\item \textbf{HCOH}: The tuple ($n_t, \eta$) is set as (1, 0.2), (1, 0.1) and (1, 0.2) on CIFAR-10, Places205 and MNIST, respectively.
\item \textbf{BSODH}: The tuple ($\lambda, \sigma, {\eta}_s, {\eta}_d$) is set as (0.6, 0.5, 1.2, 0.2), (0.3, 0.5, 1.0, 0.0) and (0.9, 0.8, 1.2, 0.2) on CIFAR-10, Places205 and MNIST, respectively.
\end{itemize}

Specific descriptions of these parameters for each method can be found in \cite{huang2013online,leng2015online,cakir2015adaptive,cakir2017online,fatih2017mihash,lin2018supervised,lin2019towards}, respectively.
Tab.\,\ref{configuration} shows the parameter settings of the proposed FCOH, which are carefully tuned and validated.
Emphatically, for SketchHash \cite{leng2015online}, the training size in each data stream has to be larger than the code length.
Following previous methods in \cite{lin2018supervised,lin2019towards}, we only report its experimental results with hashing bit being 8, 16, 32, 48.
All the experiments are run over three times and the averaged results are reported.

\begin{figure}[!t]
\begin{center}
\includegraphics[height=0.57\linewidth]{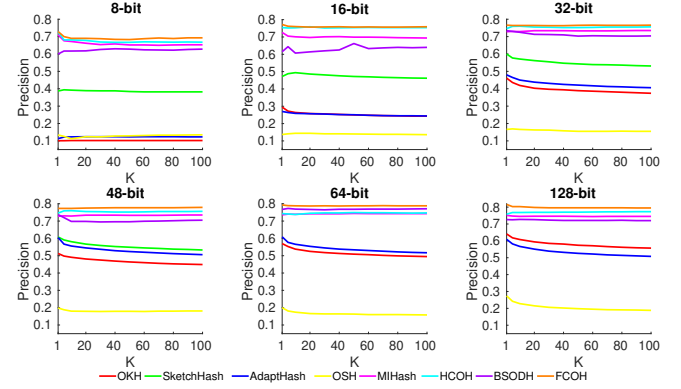}
\vspace{-2em}
\caption{\label{precision_cifar} Precision@K curves of compared algorithms on CIFAR-10.}
\end{center}
\vspace{-1.5em}
\end{figure}

\begin{figure}[th]
\begin{center}
\begin{minipage}[t]{0.48\linewidth}
\centerline{
\subfigure[\emph{m}AP]{
\includegraphics[width=\linewidth]{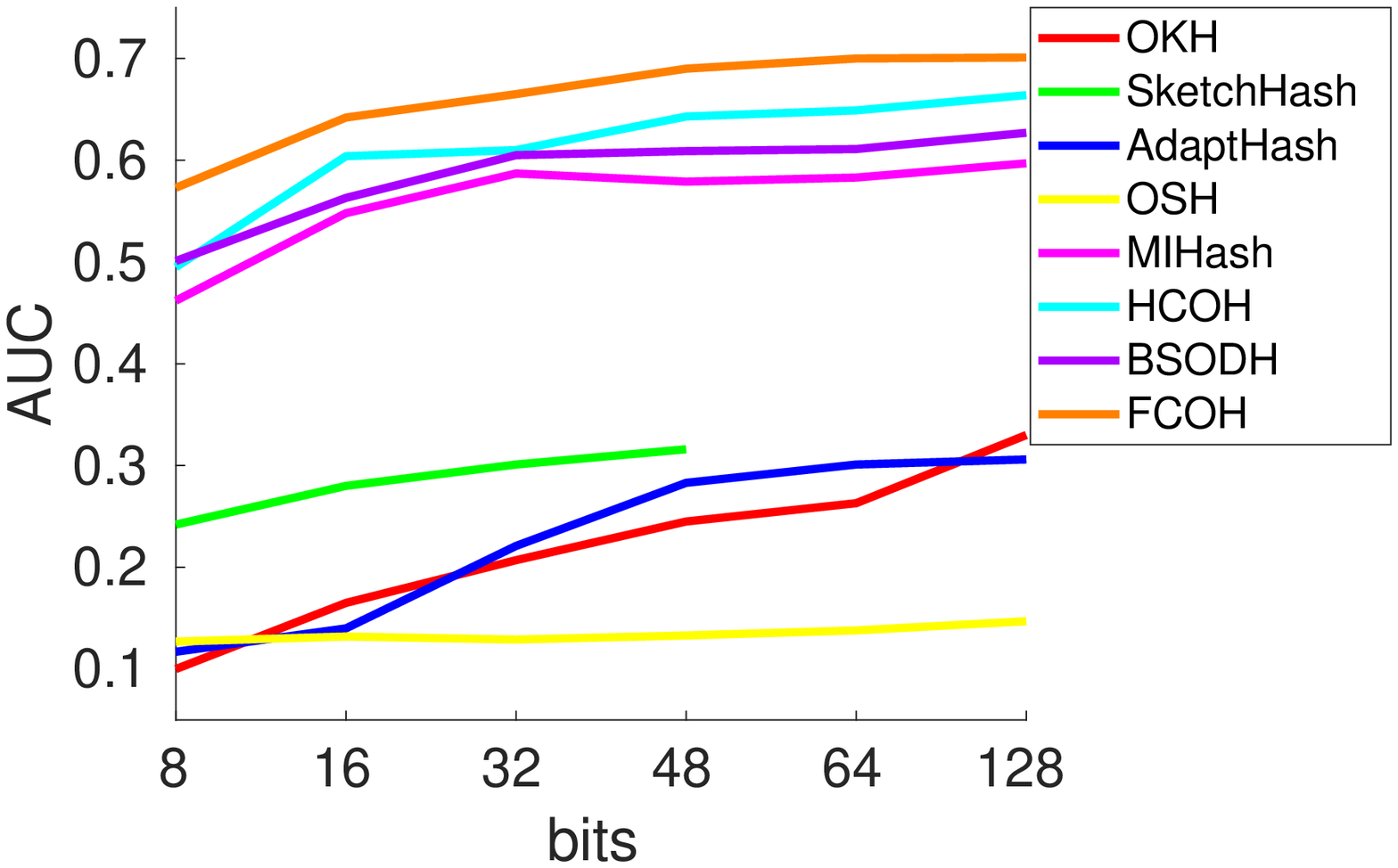}}
\subfigure[Precision@K]{
\includegraphics[width=\linewidth]{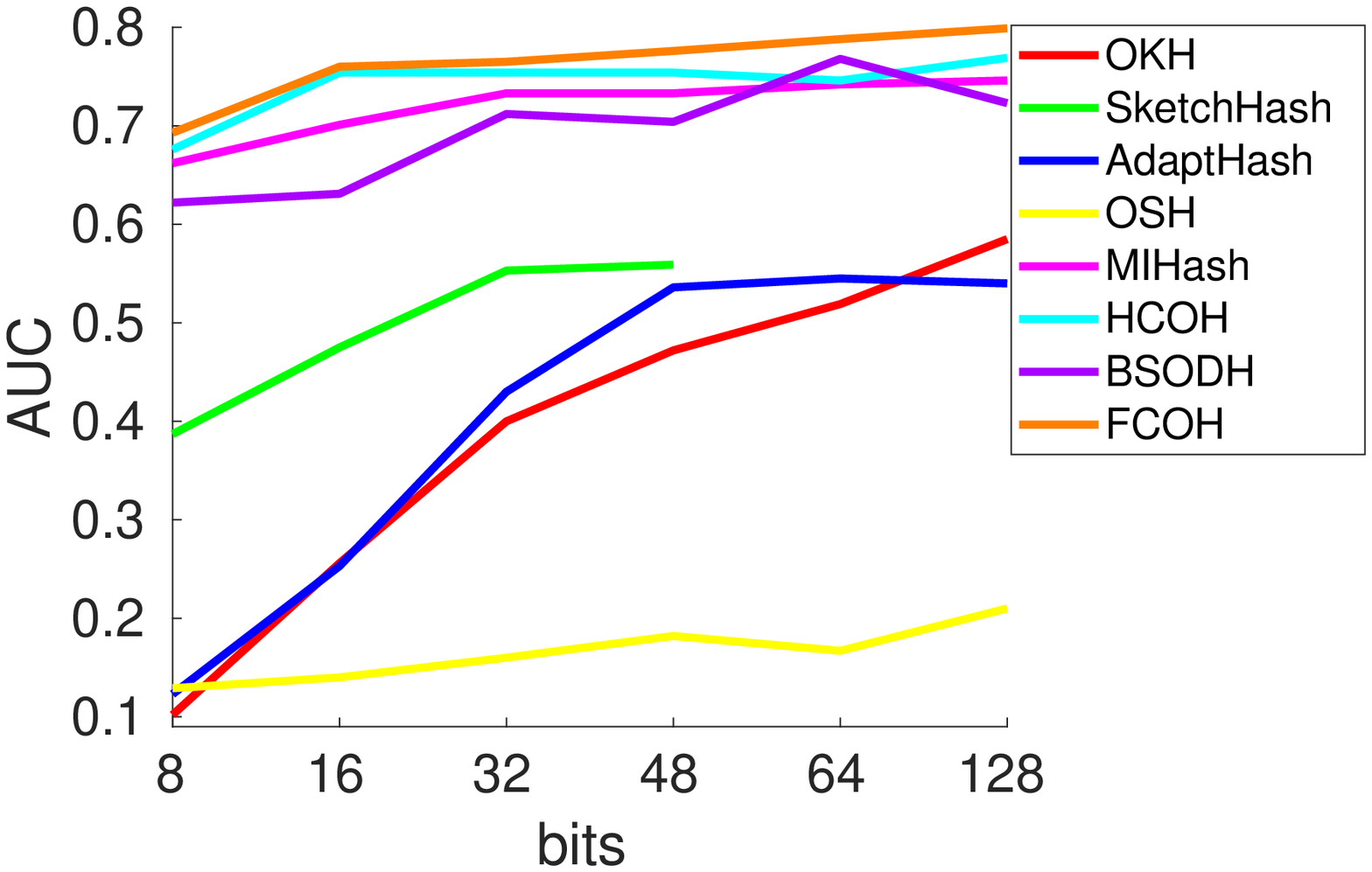}}\hspace*{0.08\linewidth}
}
\end{minipage}
\end{center}
\setlength{\abovecaptionskip}{5pt}
\vspace{-1em}
\caption{\label{auc_cifar}AUC curves for \emph{m}AP and Precision@K on CIFAR-10.}
\vspace{-1em}
\end{figure}

\subsection{Results and Discussions \label{results}}

\subsubsection{Results on CIFAR-10} \label{results_cifar}
%
%
%

%
Regarding \emph{m}AP and Precision@H2, three observations can be found:
First, as can be seen in Tab.\,\ref{map_precision_cifar}, FCOH achieves the best results in most cases.
Quantitatively, FCOH surpasses state-of-the-art methods like HCOH or BSODH by an average of $1.49\%$ \emph{m}AP and $7.94\%$ Precision@H$2$, which demonstrates the effectiveness of FCOH.
Second, for most methods, the \emph{m}AP goes up as the code length increases.
In the case of 16-bit, BSODH increases a lot, which is even better than the proposed FCOH.
Third, different from \emph{m}AP, Precision@H2 drops a lot for most methods in high bits (64 and 128).
To analyze the contrast between the second and the third observations, more information can be encoded in high bits which is beneficial to the linear scan like \emph{m}AP, while the number of distinct hash buckets to examine grows rapidly with the hash bit $r$, which is harmful to the retrieval scenario that directly returns data points within Hamming radius $2$, \emph{i.e.}, Precision@H2 \cite{cao2018deep}.

\begin{table*}[]
\centering
\caption{\emph{m}AP@1,000 And Precision@H$2$ Comparisons on Places205 with 8, 16, 32, 48, 64 And 128 bits. The Best Result Is Labeled with Boldface And the Second Best Is with An Underline.}
\vspace{-1em}
\label{map_precision_places}
\begin{tabular}{c|cccccc|cccccc}
\hline
\multirow{2}{*}{Method} & \multicolumn{6}{c|}{\emph{m}AP@1,000}                   & \multicolumn{5}{c}{Precision@H$2$}          \\
\cline{2-13}
                        & 8-bit & 16-bit & 32-bit &48-bit &64-bit & 128-bit & 8-bit & 16-bit & 32-bit &48-bit & 64-bit
                        & 128-bit \\
\hline
OKH                     &0.018  &0.033   &0.122   &0.048  &0.114  &0.258    &0.007  &0.010   &0.026   &0.017  &0.217
                        &0.075     \\
\hline
SketchHash              &0.052  &0.120   &0.202   &0.242  &  -    &  -      &0.017  &0.066   &0.220   &0.176  &  -
                        &-   \\
\hline
AdaptHash               &0.028  &0.097   &0.195   &0.223  &0.222  &0.229    &0.009  &0.051   &0.012   &0.185  &0.021
                        &0.022   \\
\hline
OSH                     &0.018  &0.021   &0.022   &0.032  &0.043  &0.164    &0.007  &0.009   &0.012   &0.023  &0.030
                        &0.059  \\
\hline
MIHash                  &\underline{0.094}&\underline{0.191}&0.244  &\underline{0.288} &0.308  &0.332    &\underline{0.022}  &\underline{0.112}   &0.204  &0.242  &0.202
                        &0.069 \\
\hline
HCOH                    &0.049  &0.173   &\underline{0.259} &0.280  &\underline{0.321}&\underline{0.347} &0.012  &0.082 &\underline{0.252} &0.179   &0.114
                        &0.036 \\
\hline
BSODH                   &0.035  &0.174   &0.250   &0.273  &0.308  &0.337    &0.009  &0.101   &0.241   &\underline{0.246} &\underline{0.212}
                        &\underline{0.101}  \\
\hline
\hline
FCOH                    &\textbf{0.099}&\textbf{0.198}&\textbf{0.267}&\textbf{0.302}&\textbf{0.335}&\textbf{0.353}
                        &\textbf{0.024}&\textbf{0.124}&\textbf{0.261}&\textbf{0.263}&\textbf{0.223}&\textbf{0.141}\\
\hline
\end{tabular}
\vspace{-1.5em}
\end{table*}

%
We further analyze the \emph{m}AP curves over different sizes of training instances and the corresponding AUC curves in Fig.\,\ref{map_instance_cifar} and Fig.\,\ref{auc_cifar}(a), which reflect our adaptivity for online learning.
Two observations can be found in Fig.\,\ref{map_instance_cifar}:
First, FCOH achieves substantially better performance at different training stages.
To take an in-depth analysis, the AUC curve for FCOH in Fig.\,\ref{auc_cifar}(a) outperforms the second best (HCOH) by a margin of $8.63\%$, which demonstrates the robustness of FCOH.
Second, FCOH obtains much better performance at the early training stages (data size is $2,000$) in different code lengths, which effectively validates the online adaptivity of the FCOH.

The results of Precision@K and their AUC curves on CIFAR-10 can be seen in Fig.\,\ref{precision_cifar} and Fig.\,\ref{auc_cifar}(b).
Generally, FCOH shows the best performance in all code lengths.
When the code length $\ge 48$, the superiority is more evident.
Regarding the AUC of Precision@K in Fig.\,\ref{auc_places}(b), FCOH transcends the second best, \emph{i.e.}, MIHash, by an average of $6.11\%$.
Notably, combing the Precision@K results in Fig.\,\ref{precision_cifar} and the \emph{m}AP results in Tab.\,\ref{map_precision_cifar}, we can conclude that, given a query, the proposed HCOH not only retrieves more relevant items, but also ranks them in the top places.

\begin{figure}[!t]
\begin{center}
\includegraphics[height=0.57\linewidth]{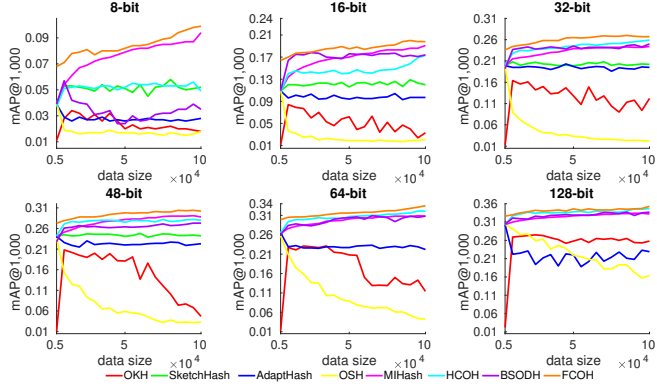}
\vspace{-2em}
\caption{\label{map_instance_places} \emph{m}AP performance with respect to different sizes of training instances on Places205.}
\end{center}
\vspace{-0.5em}
\end{figure}

\subsubsection{Results on Places205} \label{results_places}

%
Places205 is a large-scale and challenging benchmark, the results on which can well reflect the performance in real-world applications.
We start with an analysis of \emph{m}AP@$1,000$ and Precision@H$2$ in Tab.\,\ref{map_precision_places}.
As can be observed, FCOH not only ranks first but also obtains a relative increase of $3.84\%$ for \emph{m}AP@$1,000$ and $12.51\%$ for Precision@H$2$, comparing to the best baselines, HCOH or BSODH, which demonstrates the scalability of the proposed method for large-scale applications.
Moreover, the observations on CIFAR-$10$ can be also found in Places$205$, \emph{i.e.}, better \emph{m}AP and degenerated Precision@H2 in longer code length.
The explanations are the same as in Sec.\,\ref{results_cifar}, \emph{i.e.}, longer codes are beneficial to the linear scan of \emph{m}AP while harmful to the Precision@H2 which directly returns data points with Hamming radius $2$.
Besides, comparing Tab.\,\ref{map_precision_cifar} with Tab.\,\ref{map_precision_places}, for all methods, both \emph{m}AP and Precision@H2 performance on Places205 are much lower than those on CIFAR-10.
To analyze, this is due to the large scale of Places205 (millions).
It is quite challenging to obtain high performance on such a benchmark.
Nevertheless, with regard to the comparisons between different methods, FCOH yields the best in all the tests, which verifies the feasibility of FCOH in real-world applications.

\begin{figure}[!t]
\begin{center}
\includegraphics[height=0.57\linewidth]{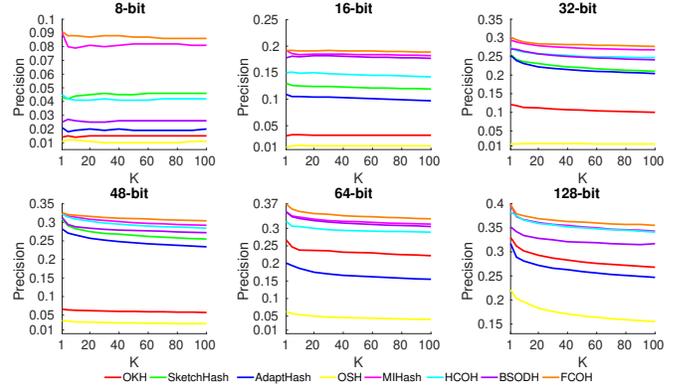}
\vspace{-2em}
\caption{\label{precision_places} Precision@K curves of compared algorithms on Places205.}
\end{center}
\vspace{-1.5em}
\end{figure}
\begin{figure}[!t]
\begin{center}
\begin{minipage}[t]{0.48\linewidth}
\centerline{
\subfigure[\emph{m}AP]{
\includegraphics[width=\linewidth]{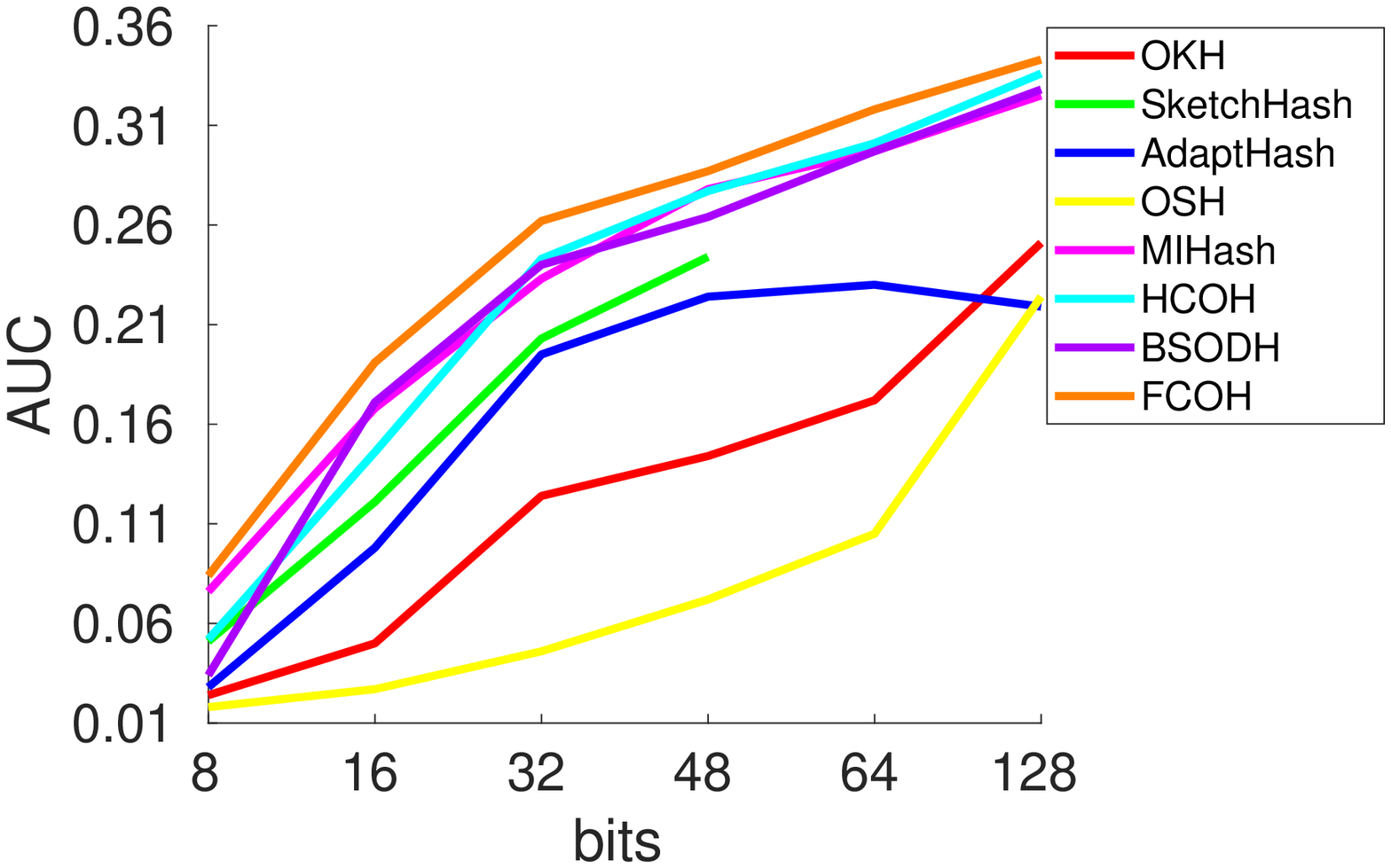}}
\subfigure[Precision@K]{
\includegraphics[width=\linewidth]{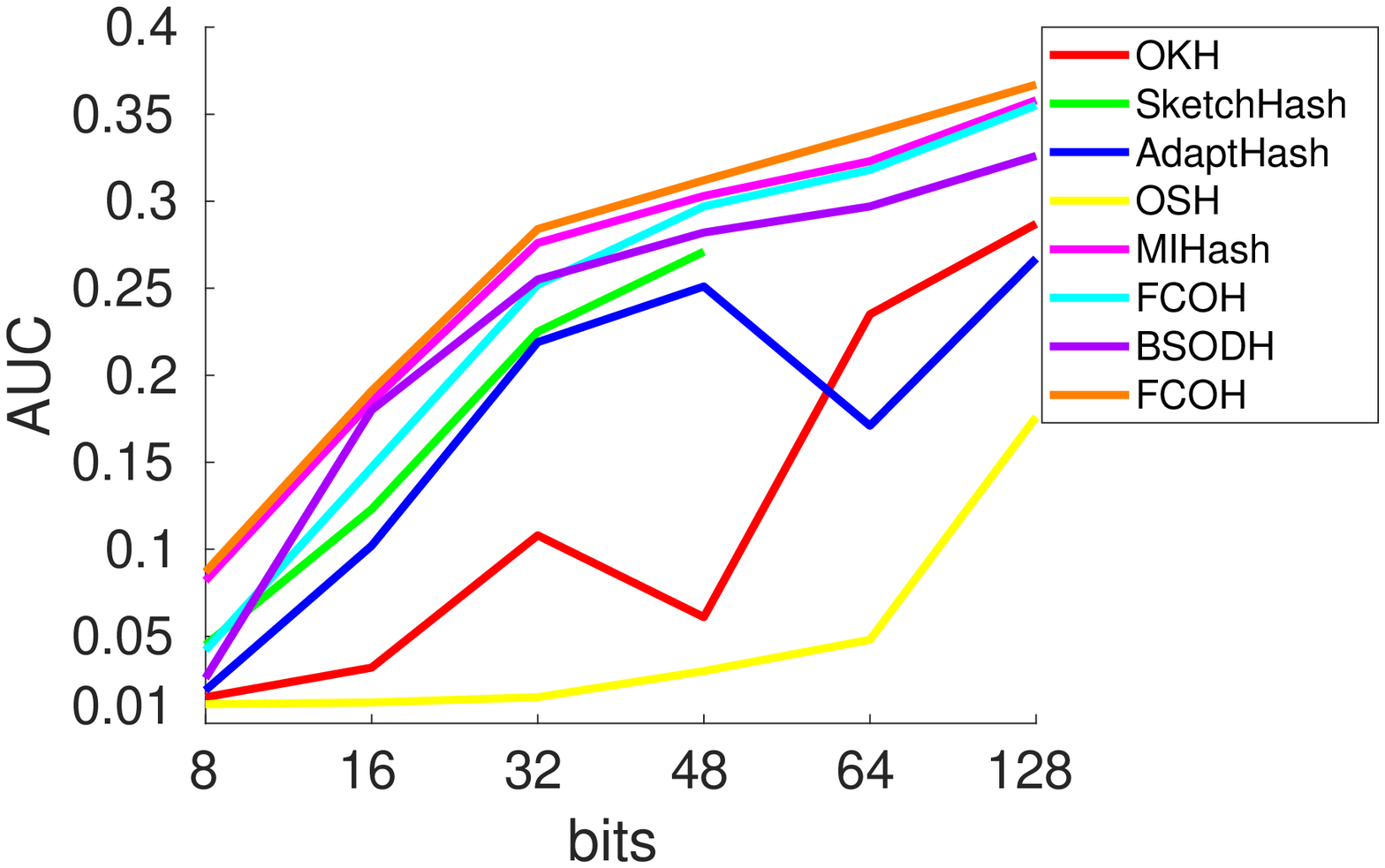}}\hspace*{0.08\linewidth}
}
\end{minipage}
\end{center}
\setlength{\abovecaptionskip}{5pt}
\vspace{-1em}
\caption{\label{auc_places}AUC curves for \emph{m}AP and Precision@K on Places205.}
\vspace{-1em}
\end{figure}

\begin{table*}[]
\centering
\caption{\emph{m}AP And Precision@H$2$ Comparisons on MNIST with 8, 16, 32, 48, 64 And 128 Bits. %
The Best Result Is Labeled with Boldface And the Second Best Is with An Underline.}
\vspace{-1em}
\label{map_precision_mnist}
\begin{tabular}{c|cccccc|cccccc}
\hline
\multirow{2}{*}{Method} & \multicolumn{6}{c|}{\emph{m}AP}                   & \multicolumn{5}{c}{Precision@H$2$}          \\
\cline{2-13}
                        & 8-bit & 16-bit & 32-bit &48-bit & 64-bit & 128-bit & 8-bit & 16-bit & 32-bit &48-bit & 64-bit
                        & 128-bit \\
\hline
OKH                     &0.100  &0.155   &0.224   &0.273  &0.301  &0.404    &0.100  &0.220   &0.457   &0.724   &0.522
                        &0.124 \\
\hline
SketchHash              &0.257  &0.312   &0.348   &0.369  &  -    &  -      &0.261  &0.596   &0.691   &0.251  & -
                        &-    \\
\hline
AdaptHash               & 0.138 &0.207   &0.319   &0.318  &0.292  &0.208    &0.153  &0.442   &0.535   &0.335  &0.163
                        &0.168  \\
\hline
OSH                     &0.130  &0.144   &0.130   &0.148  &0.146  &0.143    &0.131  &0.146   &0.192   &0.134  &0.109
                        &0.019  \\
\hline
MIHash                  &\underline{0.664} &\textbf{0.741}   &0.744   &\underline{0.780}  &0.713  &0.681 &\underline{0.487} &\underline{0.803}   &0.814  &0.739 &0.720
                        &0.471 \\
\hline
HCOH                    &0.536  &0.708 &\underline{0.756} &0.772  &0.759  &\underline{0.771}    &0.350  &0.800   &\underline{0.826}  &0.766  &0.643
                        &0.370   \\
\hline
BSODH                   &0.593  &0.700  &0.747 &0.743 &\underline{0.766}  &0.760   &0.308  &0.709   &\underline{0.826} &\underline{0.804}  &\underline{0.814}
                        &\textbf{0.643}   \\
\hline
\hline
FCOH                    &\textbf{0.673}&\underline{0.725}&\textbf{0.786}&\textbf{0.789}&\textbf{0.784}&\textbf{0.801}
                        &\textbf{0.506}&\textbf{0.817}&\textbf{0.849}&\textbf{0.814}&\textbf{0.817}&\underline{0.620}\\
\hline
\end{tabular}
\vspace{-1.5em}
\end{table*}

Given the results of \emph{m}AP@$1,000$ in Fig.\,\ref{map_instance_places}, we can see that FCOH performs best in all cases.
It demonstrates the robustness of FCOH when the training proceeds as reflected in Fig.\,\ref{auc_places}(a), where FCOH gains an average of $18.14\%$ AUC improvements compared to HCOH.
Besides, it also shows the fast online adaptivity of FCOH since the proposed method grows fast and outperforms other methods by large margins at the early training stages (data size 5,000).
Note that when the hashing bits range from $8$ to $64$, FCOH shows a better margin than other methods.
However, in the case of $128$-bit, FCOH and BSODH show a similar performance (FCOH is slightly better.).
To explain, as mentioned in Sec.\,\ref{results_cifar}, the \emph{m}AP increases with the increase of the code length.
When the code length is $128$, other methods can also obtain relatively high results.
Nevertheless, in all code lengths, the proposed method consistently shows the best performance, which further demonstrates the superiority of FCOH.

Lastly, the Precision@K performance in Fig.\,\ref{precision_places} and their AUC results Fig.\,\ref{auc_places}(b) demonstrate again the effectiveness of FCOH.
As can be observed, on this large-scale and challenging dataset, FCOH obtains higher precision results in Fig.\,\ref{precision_places} and transcends the state-of-the-art MIHash with an AUC gain of $3.78\%$.
Together with the \emph{m}AP@$1,000$ performance in Fig.\,\ref{map_instance_places}, we can conclude that, given a query, FCOH not only retrieves more relevant items but ranks them in the top position, which highly meets the need of real-world applications.

\begin{figure}[!t]
\begin{center}
\includegraphics[height=0.57\linewidth]{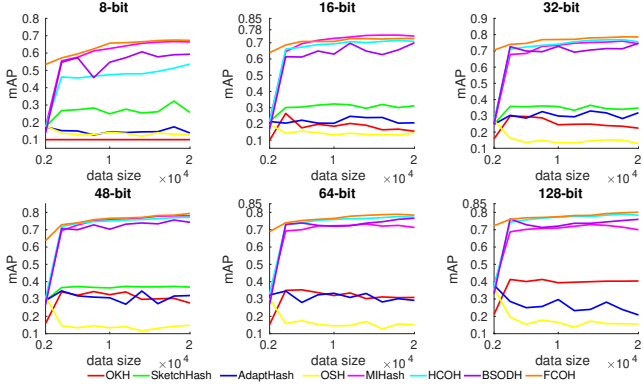}
\vspace{-2em}
\caption{\label{map_instance_mnist} \emph{m}AP performance with respect to different sizes of training instances on MNIST.}
\end{center}
\vspace{-0.5em}
\end{figure}

\subsubsection{Results on MNIST}\label{results_mnist}
%
%
%
%

%
In Tab.\,\ref{map_precision_mnist}, the proposed method shows the best results except for the cases of \emph{m}AP in 16-bit and Precision@H2 in 128-bit, where FCOH ranks the second.
The experimental results in Tab.\,\ref{map_precision_mnist} (on MNIST) are similar to those in Tab.\,\ref{map_precision_cifar} (on CIFAR-10).
On one hand, the quantitative values on MNIST and CIFAR-10 are much larger than those on Places205.
On the other hand, regarding \emph{m}AP in $16$-bit, MIHash obtains the best on both MNIST and CIFAR-10. In terms of Precision@H2 in high bits, BSODH obtains the best on both MNIST (64-bit) and CIFAR-10 (128-bit).
This is because MNIST and CIFAR-10 have the similar quantitative scale ($60$K $\sim$ $70$K) and categories ($10$ for both), which are two relatively simple benchmarks compared with the scale of Places205 (millions of samples and $205$ categories).
Nevertheless, FCOH still yields an overall better performance.
Generally speaking,
the proposed FCOH achieves state-of-the-art results with an average increase of $1.76\%$ for \emph{m}AP and $1.08\%$ for Precision@H$2$  comparing to the second best.
Hence, the superiorities of FCOH are demonstrated.
%

\begin{figure}[!t]
\begin{center}
\includegraphics[height=0.57\linewidth]{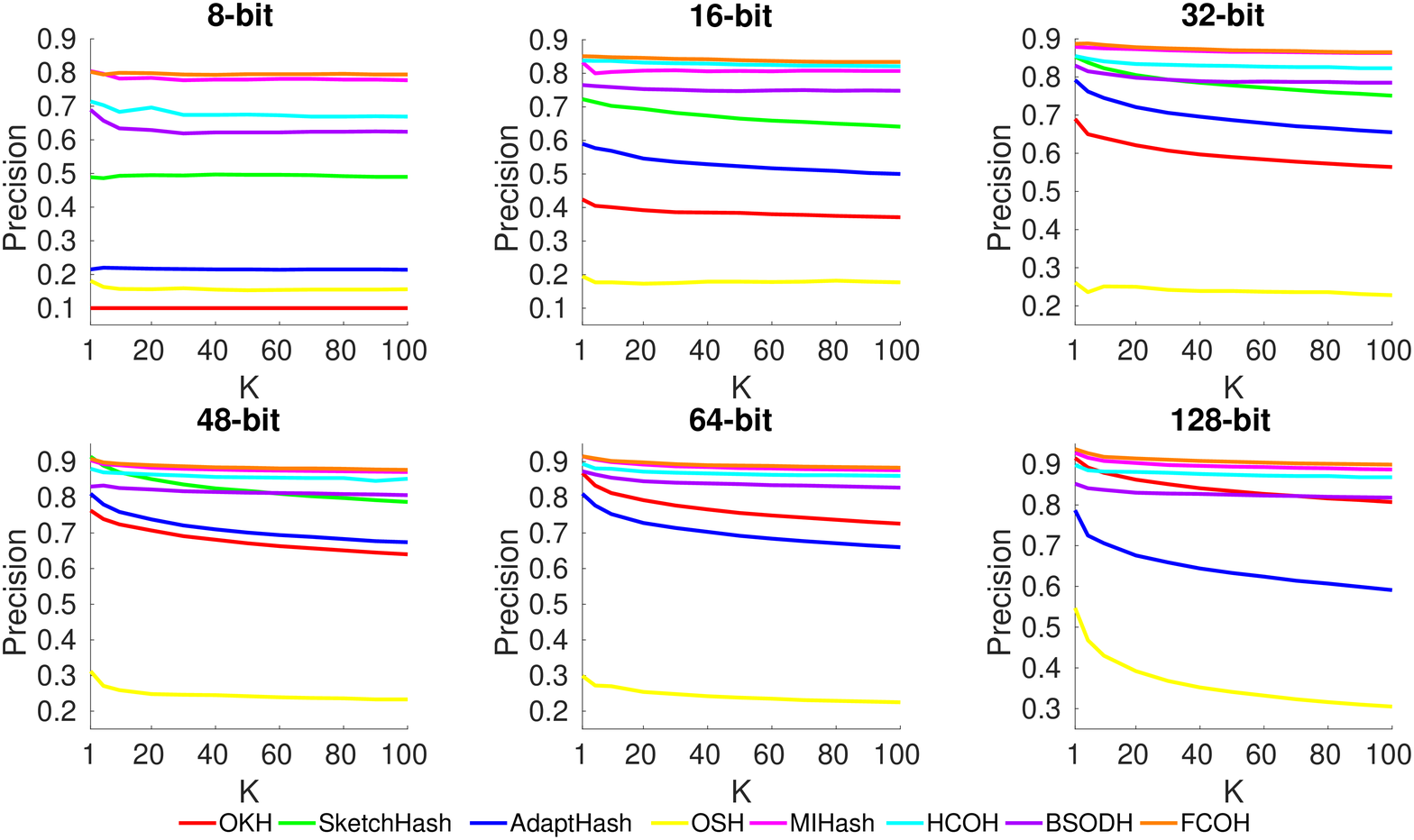}
\vspace{-2em}
\caption{\label{precision_mnist} Precision@K curves of compared algorithms on MNIST.}
\end{center}
\vspace{-1.5em}
\end{figure}

\begin{figure}[th]
\begin{center}
\begin{minipage}[t]{0.48\linewidth}
\centerline{
\subfigure[\emph{m}AP]{
\includegraphics[width=\linewidth]{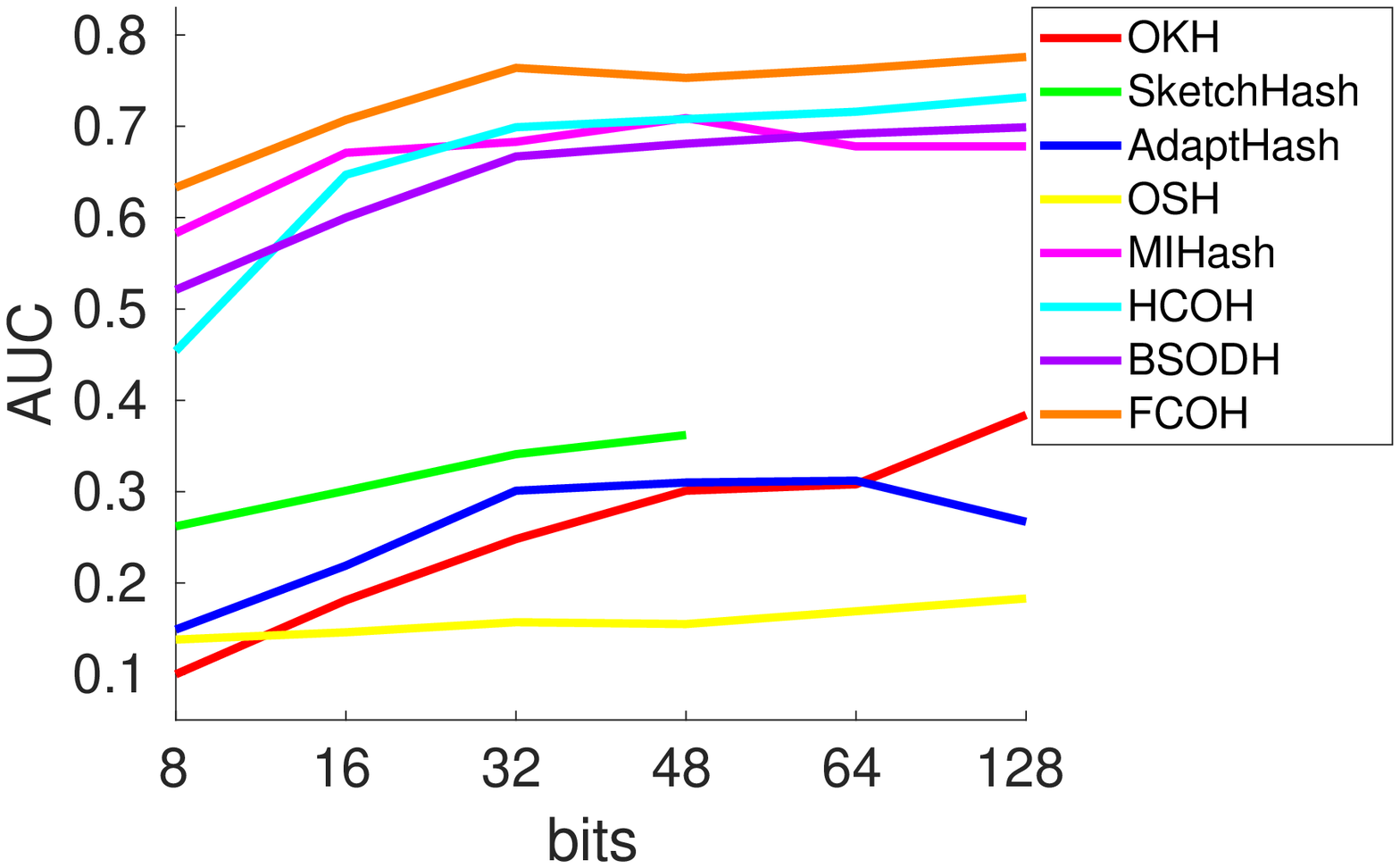}}
\subfigure[Precision@K]{
\includegraphics[width=\linewidth]{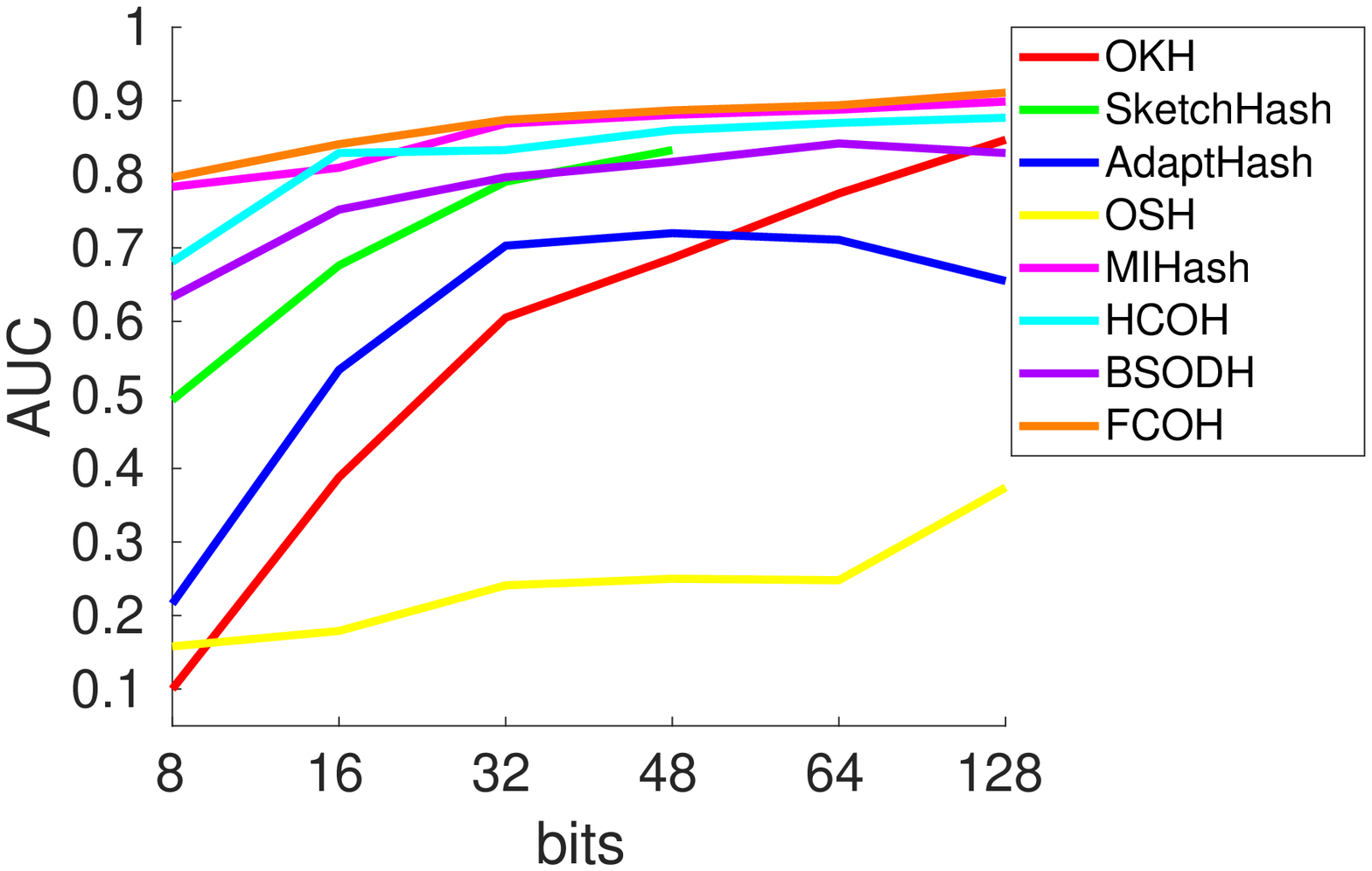}}\hspace*{0.08\linewidth}
}
\end{minipage}
\end{center}
\setlength{\abovecaptionskip}{5pt}
\vspace{-1em}
\caption{\label{auc_mnist}AUC curves for \emph{m}AP and Precision@K on MNIST.}
\vspace{-1em}
\end{figure}

\begin{figure*}
\begin{center}
\centerline{
\includegraphics[height=0.18\linewidth]{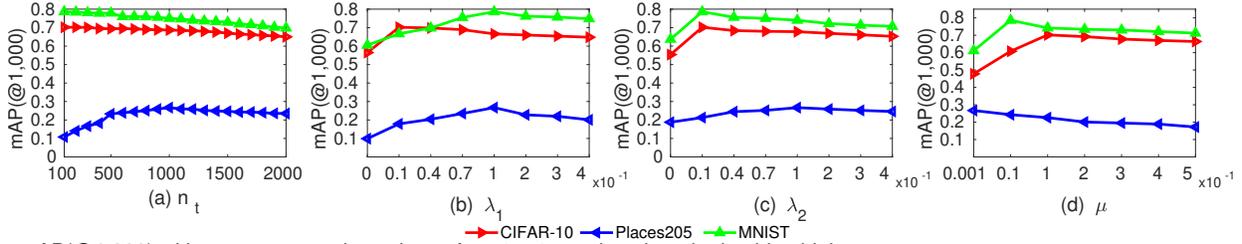}}
\vspace{-1.5em}
\caption{\label{all_parameters} \emph{m}AP(@1,000) with respect to varying values of $n_t$, ${\lambda}_1$, ${\lambda}_2$ and $\mu$ when the hashing bit is $32$.}
\end{center}
\vspace{-2.5em}
\end{figure*}

Besides, we argue that Fig.\,\ref{map_instance_mnist} and Fig.\,\ref{auc_mnist}(a) also prove the online adaptivity of FCOH.
First, FCOH keeps consistently better \emph{m}AP in different sizes of training instances and obtains an increase of $9.83\%$ AUC compared with the best baseline HCOH.
On the other hand, FCOH transcends other methods in early training stages by a large margin.
Taking the hash bit of $64$ as an instance, FCOH gets an \emph{m}AP of $0.689$ while it is only $0.323$ for MIHash, $0.329$ for HCOH and $0.270$ for BSODH when the training size is $2,000$.
Hence, the fast adaptivity ability of FCOH is well demonstrated.

As for Precision@K, FCOH is still competitive in all cases with different code lengths, as reflected in Fig.\,\ref{precision_mnist}.
Quantitatively speaking, FCOH obtains an AUC increase of $1.48\%$ compared with the second best of MIHash.
Besides, different from the results on CIFAR-10 and Places205, the increments on MNIST are small.
To analyze, MNIST is a simple benchmark and it is easy for existing methods to obtain high performance.
When the code length $\le 16$, existing methods can achieve an accuracy of more than $80\%$.
When the code length $\ge 32$, existing methods can obtain a high accuracy of near $90\%$.
Hence, it is tough to further obtain a significant improvement upon MNIST.

\subsubsection{In-depth Analysis on Better Performance}
As discussed in Sec.\,\ref{results_cifar}, Sec\,\ref{results_places} and Sec.\,\ref{results_mnist}, FCOH shows an overall better performance than the SOTAs. To analyze, class-wise updating focuses on minimizing the intra-class distance and pulls other classes away from the current optimized class, leading to a fast adaptivity to the new data stream as shown in Sec.\,\ref{class_wise_updating}. Semi-relaxation optimization treats part of the binary constraints as a constant, which is computed by hash weights learned from the last round. Thus, it can update the hash model by using the new data stream, meanwhile preserving the past information as shown in Sec.\,\ref{semi_relaxation_optimization}. Therefore, the better performance of FCOH is the collective effort of both class-wise updating and semi-relaxation optimization.

\begin{table}[!t]
\centering
\caption{\label{variants}\emph{m}AP(@1,000) Results of 32-Bit for Different Variants of the Proposed Method on the Three Datasets.}
\vspace{-1em}
\begin{tabular}{c|c|c|c}
\hline
        & CIFAR-10 & Places205 & MNIST \\ \hline
BSODH (baseline) & 0.689 &0.250 &0.747 \\ \hline
FCOH$_S$ & 0.691    & 0.255     & 0.760 \\ \hline
FCOH$_C$ & 0.696    & 0.262     & 0.777 \\ \hline
FCOH    & \textbf{0.702} &\textbf{0.267} &\textbf{0.786} \\ \hline
\end{tabular}
\vspace{-2em}
\end{table}

By removing the class-wise updating and semi-relaxation, our FCOH is the same with BSODH \cite{lin2019towards}, which can serve as the baseline to show the effectiveness of the proposed class-wise updating and semi-relaxation.
To that effect, two variants of FCOH are introduced. 
The first variant is denoted as FCOH$_S$ where data from all classes are involved in the training at each round and the proposed semi-relaxation is kept.
The second variant is denoted as FCOH$_C$ where the class-wise updating is kept and the semi-relaxation is discarded. Instead, we apply the discrete cyclic coordinate descent (DCC) developed in \cite{shen2015supervised}, as mentioned in Sec.\,\ref{semi_relaxation_optimization}.
DCC is also the standard optimizer of BSODH \cite{lin2019towards}.
Then, we take 32-bit as an example, and show the mAP(@1,000) performance on the three datasets. The experimental results are listed in Tab.\,\ref{variants}.

As can be seen, FCOH, which contains both class-wise updating and semi-relaxation optimization, shows the optimal results, and its variants, \emph{i.e.}, FCOH$_S$ and FCOH$_C$ obtain less mAP(@1,000) performance. Among all, BSODH without both class-wise updating and semi-relaxation optimization results in the least performance.
Thus, it well verifies the analysis above that the better results of FCOH are the collective effort of the proposed class-wise updating and semi-relaxation optimization.

\subsubsection{Convergence of the Class-wise Updating}

\begin{figure}[!t]
\begin{center}
\includegraphics[height=0.5\linewidth]{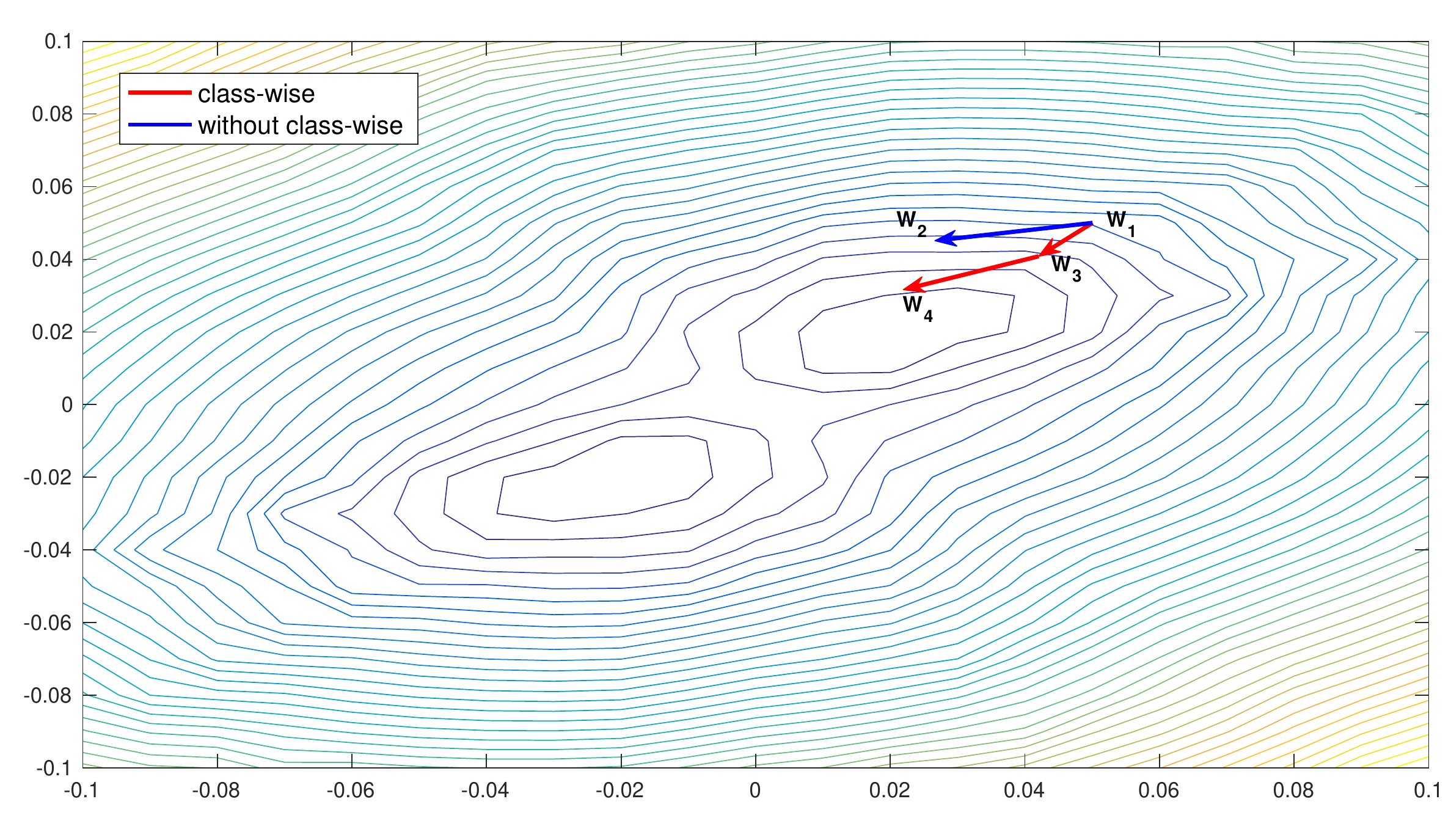}
\vspace{-1em}
\caption{\label{class_wise_gradient2}
Visualization with/without the class-wise updating.
$\mathbf{W}_1$ is the start point.
Without class-wise updating, the renewed weights come at $\mathbf{W}_2$, 
With class-wise updating, the hash weights are updated gradient-by-gradient.
It first updates hash functions using data from the first class, which returns $\mathbf{W}_3$, based on which the second-class data is used and then $\mathbf{W}_{4}$ is obtained. As can be seen, the final $\mathbf{W}_{4}$ rewards a better solution than $\mathbf{W}_2$ without class-wise updating.
}
\end{center}
\vspace{-2em}
\end{figure}

In Fig.\,\ref{class_wise_gradient}, we illustrate a toy example which shows that the class-wise updating can obtain a better solution. In this section, we visualize the updating of Eq.\,\ref{imbalance} with/without class-wise updating in Fig.\,\ref{class_wise_gradient2} for a rigorous demonstration \footnote{The figure is plotted using contour function with MATLAB: https://www.mathworks.com/help/matlab/ref/contour.html.}. We implement Fig.\,\ref{class_wise_gradient2} with two-class synthetic data and the hash codes are updated using the DCC \cite{shen2015supervised} to exclude the effect of semi-relaxation optimization. It can be seen that updating with a class-wise fashion reaches a more optimal solution than the one without class-wise updating, which shows that class-wise updating provides a more effective facility for online hash learning.

\subsection{Ablation Study}
In this subsection, we conduct the ablation studies on the hyper-parameters of FCOH, including the batch size of streaming data, \emph{i.e.}, $n_t$, as defined in Sec.\,\ref{definition};
the balance parameters, \emph{i.e.}, ${\lambda}_1$ and ${\lambda}_2$, as defined in Eq.\,(\ref{semi_relax});
and the learning rate $\mu$, as defined in Eq.\,(\ref{optimization}).
Without loss of generality, we conduct experiments with varying values of these hyper-parameters with respect to \emph{m}AP (\emph{m}AP@1,000) in the case of $32$-bit in Fig.\,\ref{all_parameters} (Detailed values used in this paper is outlined in Tab.\,\ref{configuration}.).
Similar experimental results can be observed in other hashing bits.

\subsubsection{The Influence of $n_t$} \label{n_t}
We first start with the analysis on the varying values of batch size $n_t$ at the $t$-stage.
The experiments are sampled with $n_t$ ranged in $\{ 100, 200, ..., 2,000 \}$ and the results are shown in Fig.\,\ref{all_parameters}(a).
As can be observed, on CIFAR-10 and MNIST, the \emph{m}AP(\emph{m}AP@1,000) degenerates along with the increase of batch size $n_t$,
while on Places205,  the \emph{m}AP(\emph{m}AP@1,000) achieves the optimal when $n_t \approx 1,000$.
In the experiments, we empirically set the values of $n_t$ as $100$ on CIFAR-10 and MNIST, and $1,000$ on Places205 as stated in Tab.\,\ref{configuration}.
Such settings conform with the requirements of online learning since $n_t \ll n$ in our experiments ($n = 20K$ on CIFAR-10 and MNIST, and $n = 100K$ on Places205).

We observe that for CIFAR-10 and MNIST, they have similar data size and share the same batch size. As for Places205, it has a much larger data size than CIFAR-10 and MNIST, and the required batch size is also larger than the others. To explain, Places205 is a large-scale retrieval set. Setting a smaller value makes it hard to capture and adapt to the data property. The three datasets represent different scenes in real life. These different scenes typically require different configurations to deploy models. Hence, it’s intuitive that different datasets have different optimal $n_t$ values.

\subsubsection{The Influence of ${\lambda}_1$} \label{lambda_1}
As shown in Eq.\,(\ref{final}), ${\lambda}_1$ is used to reflect the importance of similar items in the inner product based learning scheme.
Fig.\,\ref{all_parameters}(b) plots the influence of different values of ${\lambda}_1$ to the performance.
Generally speaking, when ${\lambda}_1 = 0.01$ on CIFAR-10, ${\lambda}_1 = 0.1$ on Places205 and MNIST, FCOH obtains the best \emph{m}AP (\emph{m}AP@1,000) on the three benchmarks ($0.702$ on CIFAR-10, $0.267$ on Places205 and $0.786$ on MNIST).
Besides, when ${\lambda}_1 = 0$, FCOH suffers a great performance loss as can be seen in Fig.\,\ref{all_parameters}(b).
More specifically, in this case, the \emph{m}AP (\emph{m}AP@1,000) scores are $0.565$, $0.099$ and $0.605$ on CIFAR-10, Places205 and MNIST, respectively.
To analyze, when ${\lambda}_1 = 0$, FCOH only relies on dissimilar items to learn hash functions.
The to-be-learned binary codes will be separable from each other.
Therefore, the performance degenerates a lot.
In the experiments, we empirically set the values of ${\lambda}_1$ as $0.01$ on CIFAR-10, and $0.1$ on Places205 and MNIST.

\begin{table*}[!t]
\centering
\caption{Time Consumption on Hash Function Updating and Hash Table Updating on the Three Benchmarks under $32$-Bit Hashing Codes. \label{training_time}}
\vspace{-1em}
\begin{tabular}{c|c|c|c|c|c|c|c|c|c}
\hline
\multirow{2}{*}{Method} & \multicolumn{3}{c|}{CIFAR-10 (s)}    & \multicolumn{3}{c|}{Places205 (s)}    & \multicolumn{3}{c}{MNIST (s)}          \\ \cline{2-10} 
                        & Hash Function & Hash Table & Total & Hash Function & Hash Table & Total & Hash Function & Hash Table & Total \\ \hline
OKH                     & 4.53          & 1,600       &1,604.53  & 15.66   & 12,500    &12,515.66  & 4.58    & 400     &404.58      \\ \hline
SketchHash              & 4.98          & 64         &68.98     & 3.52    & 500       &503.52     & 1.27    & 16      &17.27    \\ \hline
AdaptHash               & 20.73         & 1,600      &1,620.73  & 14.49   & 12,500    &12,514.49  & 6.26    & 400     &406.26      \\ \hline
OSH                     & 93.45         & 3,200      &3,293.45  & 56.68   & 25,000    &25,056.68  & 24.07   & 800     &824.07      \\ \hline
MIHash                  & 120.10        &103.20     &223.30     & 468.77  & 437.25    &906.02  & 97.59   &24.52  &122.11  \\ \hline
HCOH                    & 12.34         & 3,200      &3,212.34  & 10.54   & 25,000    &25,010.54  & 4.01    & 800     &804.01      \\ \hline
BSODH                   & 36.12         & 1.60       &37.72     & 69.73   & 5         &74.73   & 4.83       & 0.40    &5.23      \\ \hline
FCOH                    & 7.81          & 32         &39.81     & 15.27   & 25        &40.27    & 1.64      & 8       &9.64      \\ \hline
\end{tabular}
\vspace{-2em}
\end{table*}


\subsubsection{The Influence of ${\lambda}_2$}  \label{lambda_2}
As shown in Eq.\,(\ref{final}), ${\lambda}_2$ is used to reflect the importance of dissimilar items in the inner product based learning scheme.
From Fig.\,\ref{all_parameters}(c), we can observe that when ${\lambda}_2 = 0.01$ on CIFAR-10 and MNIST, ${\lambda}_2 = 0.1$ on Places205, FCOH obtains the best \emph{m}AP (\emph{m}AP@1,000) of $0.702$, $0.267$ and $0.786$, respectively.
Similar to ${\lambda}_1$, when ${\lambda}_2 = 0$, FCOH also suffers a great performance loss ($0.554$ on CIFAR-10, $0.188$ on Places205 and $0.637$ on MNIST).
However, the explanation behind this phenomenon is different from the case of ${\lambda}_1 = 0$.
Specifically, when ${\lambda}_2 = 0$, the proposed FCOH learns hash functions simply through similar items.
Under this situation, the learned binary codes will be as close as possible, which inevitably causes the performance degeneration.
In the experiments, we empirically set the values of ${\lambda}_2$ as $0.01$ on CIFAR-10 and MNIST, and $0.1$ on Places205.

Combining Sec.\,\ref{lambda_1} and Sec.\,\ref{lambda_2} together, the optimal values for the tuple (${\lambda}_1, {\lambda}_2$) are ($0.01, 0.01$) on CIFAR-10, ($0.1, 0.1$) on Places205, and ($0.1, 0.01$) on MNIST, respectively.
Note that, the tuple set above shows the equal (or approximate) importance between similar items and dissimilar items since ${\lambda}_1$ and ${\lambda}_2$ share the same (or similar) values on the three benchmarks.
That is to say, the proposed class-wise updating strategy can well solve the problem of ``data imbalance" since both similar information and dissimilar information are equally learned in our framework\footnote{To analyze, the proposed class-wise updating method reduces the quantitative difference between similar items and dissimilar items each time when updating the hash functions.
For example, there have 10 similar items and 100 dissimilar items.
Through class-wise updating, there may have 2 similar items and 10 dissimilar items when updating the $c$-th category.}.

\subsubsection{The Influence of $\mu$}   \label{mu}
Lastly, the influence of learning rate $\mu$ is shown in Fig.\,\ref{all_parameters}(d).
Generally, \emph{m}AP(\emph{m}AP@1,000) is sensitive to the varying $\mu$.
To obtain the best \emph{m}AP(\emph{m}AP@1,000), in the experiments, we empirically set the values of $\mu$ as $0.01$, $0.0001$ and $0.1$ on CIFAR-10, Places205 and MNIST, respectively.

Besides the observation that ${\lambda}_1 \approx {\lambda}_2$ as mentioned in Sec.\,\ref{lambda_2},
we empirically find that the suitable value of ${\lambda}_1$, ${\lambda}_2$ and $\mu$ fall in $10^{-m}$, where $m=1,2,3,4,5$.
Hence, in practical applications, the users can try these different values and choose the best one.

\subsection{Training Efficiency} \label{training_efficiency}
To quantitatively evaluate the training efficiency of FCOH, including the updating of hash functions and hash table,
we conduct experiments given hashing bit $r = 32$ in Tab.\,\ref{training_time}.
Similar observations can also be found in other code lengths.
\textbf{Hash Function Updating}.
Generally, OKH and SketchHash are the most efficient ones, which however suffer poor \emph{m}AP (\emph{m}AP@1,000).
Compared with state-of-the-art methods (MIHash and BSODH), FCOH shows consistently better efficiency by a margin.
As analyzed in Sec.\,\ref{introduction}, MIHash has to calculate the Hamming distance between the neighbors and non-neighbors for each instance, while the discrete optimization adopted in BSODH brings about more variables and has the convergence problem.
Compared with HCOH, the proposed method still costs less training time on CIFAR-10 and MNIST, while on Places205 HCOH is more efficient.
To analyze, FCOH asks for a larger batch size on Places205 ($n_t = 2,000$ in Tab.\,\ref{configuration}.) while $n_t =1$ for HCOH.
Nevertheless, HCOH suffers poor performance especially in low hashing bits in the experiments.
It can be observed that FCOH costs more training time on CIFAR-10 and Places205 than on MNIST.
We argue that this is owing to the high feature dimension of CIFAR-10 ($4,096$-dim) and large scale of Places205 ($100,000$).

\textbf{Hash Table Updating}.
We further compare the time consumption on hash table updating.
The hash table needs to be renewed every time when updating the hash functions.
%
%
It is easy to know that the total times spent on the hash table updating depends on the times of updating hash functions, \emph{i.e.}, $\frac{n}{n_t}$, where $n$ is the size of the dataset and $n_t$ is the size of image batch at each updating stage.
From Tab.\,\ref{training_time}, it can be seen that OKH and HCOH are most inefficient in hash table updating.
This is because the batch size for them is 1, which means more updating of hash table is required.
Note that, MIHash also requires $n_t = 1$, however, it consumes less time in hash table updating.
To analyze, MIHash does not update hash table each time the hash functions are updated.
Instead, it devises a ``gating” mechanism where the hash table is updated only when the mutual information of updated hash model is better than the currently best hash model.
Hence, MIHash requires fewer times of hash table updating.
Besides, we observe that BSODH requires the least time of hash table updating.
Taking a deeper analysis, BSODH has the largest batch size ($n_t$ = 2,000 for CIFAR-10 and MNIST, and 5,000 for Places205).
Lastly, FCOH also merits in its efficient hash table updating as reflected in Tab.\,\ref{training_time}, especially better than the gating-designed MIHash.
Though BSODH is more efficient in hash table updating than FCOH, the large batch size makes it inefficient in hash function updating.

Above all, the proposed method is efficient in both hash function updating and hash table updating.
Notably, the main focus of this paper is to design an efficient and effective method for hash function updating.
Nevertheless, some existing works \cite{ma2017partial,weng2020online} which are particularly designed for hash table updating can be integrated into the above online hashing methods to further improve the efficiency of hash table updating.

\subsection{Comparisons with offline Hashing Method}
\begin{figure}[!t]
\begin{center}
\begin{minipage}[t]{0.45\linewidth}
\centerline{
\subfigure[\emph{m}AP Performance]{
\includegraphics[width=\linewidth]{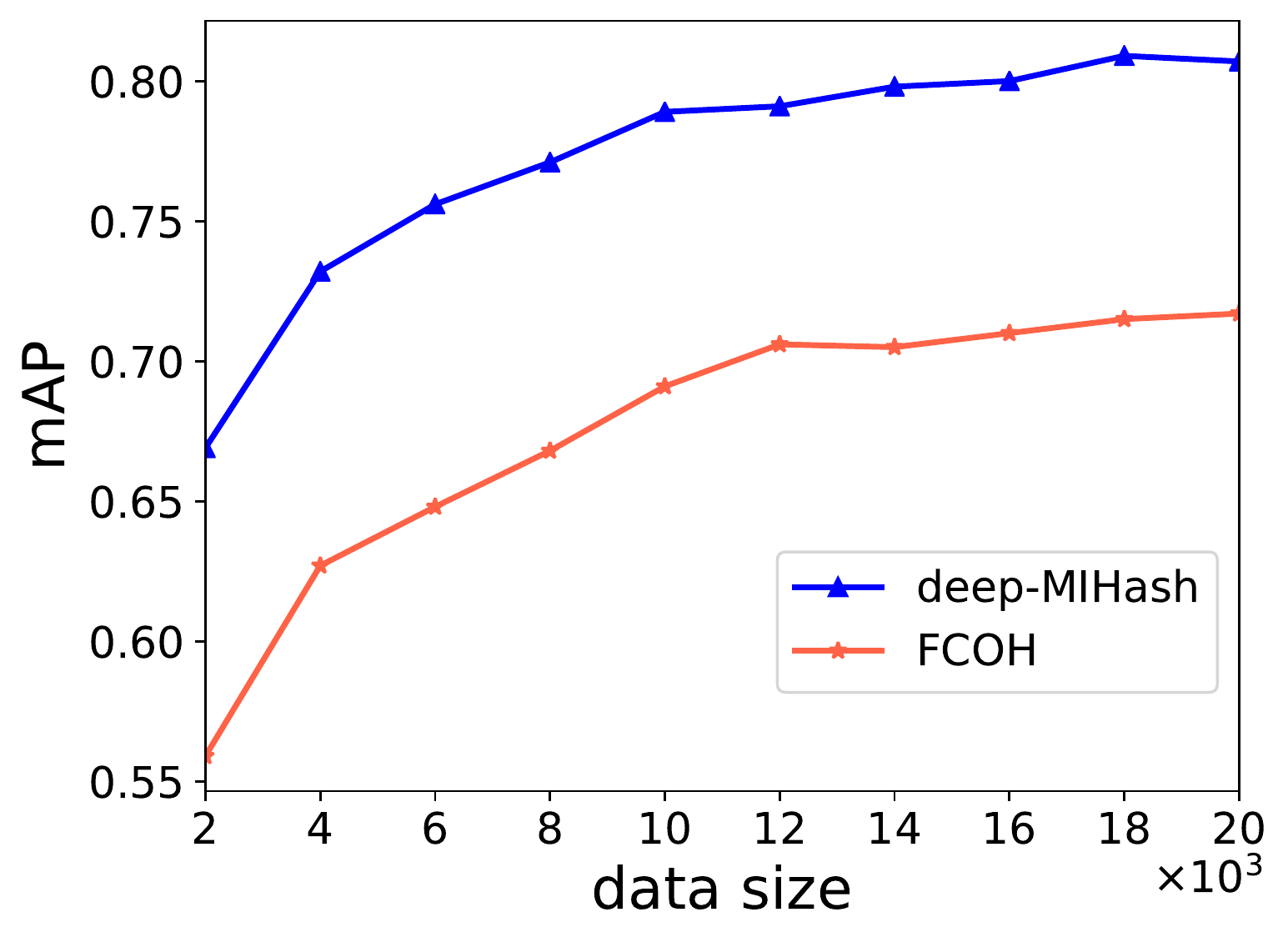}}
\subfigure[Training Time]{
\includegraphics[width=\linewidth]{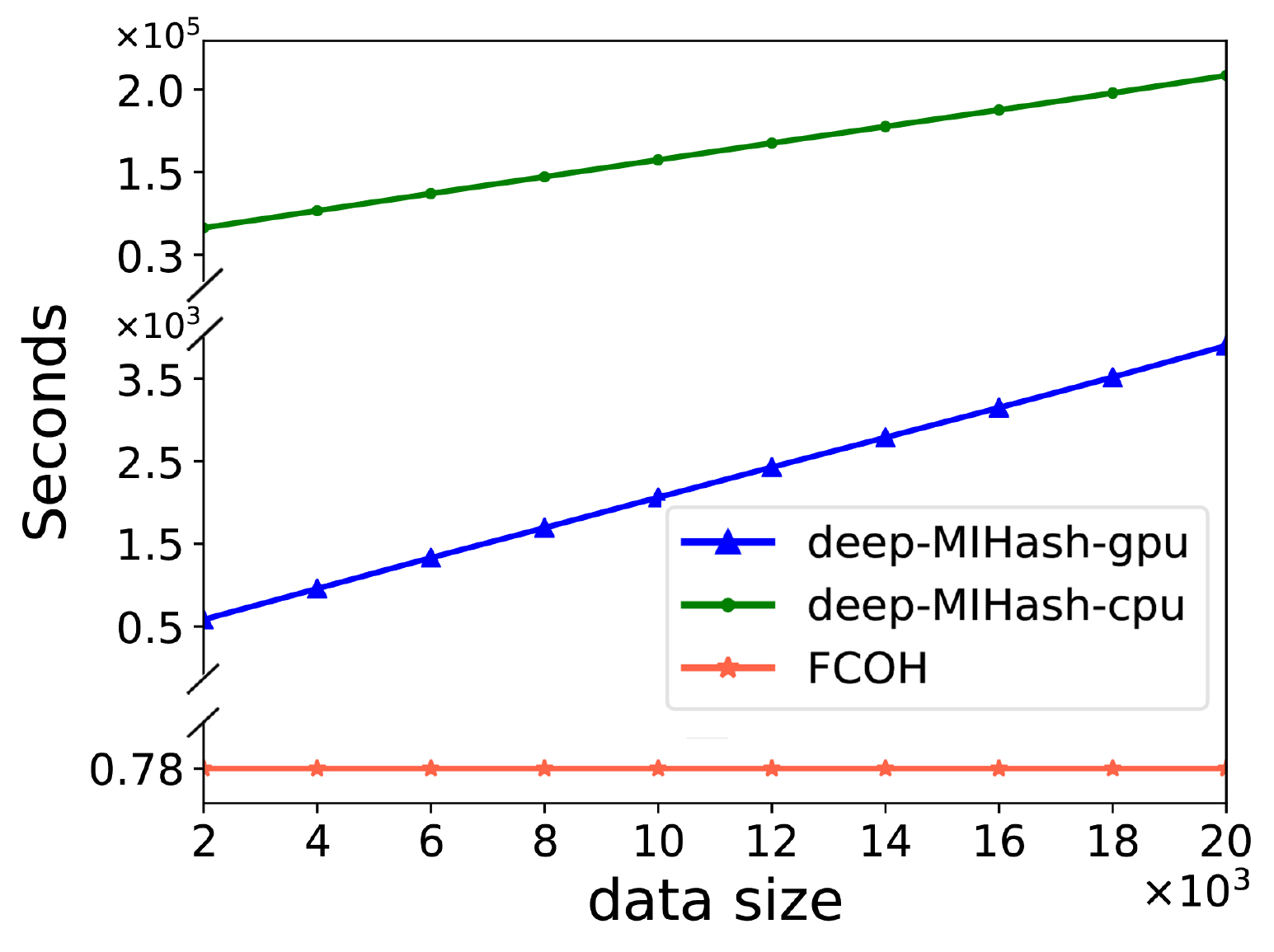}}\hspace*{0.08\linewidth}
}
\end{minipage}
\end{center}
\setlength{\abovecaptionskip}{5pt}
\vspace{-1em}
\caption{\label{offline}\emph{m}AP performance and time consumption on hash function updating at each training stage on CIFAR-10.}
\vspace{-1em}
\end{figure}

The main focus of online hashing is to efficiently deal with the streaming data.
Traditional offline hashing could also do this job by re-training the model with the accumulated data, which however has extremely heavy training consumption, but merits in better accuracy.
In this subsection, we further conduct experiments to compare with the state-of-the-art deep-MIHash \cite{cakir2019hashing}, which is an offline extension of MIHash \cite{fatih2017mihash}.
To that effect, we conduct experiments in 32-bit on CIFAR-10, and analyze the \emph{m}AP performance and their training time, the results of which are shown in Fig.\,\ref{offline}.
For deep-MIHash, we display its training time with both GPU and CPU.
As can be seen, deep-MIHash obtains better \emph{m}AP performance, \emph{i.e.}, around 10\% improvements over FCOH at each updating stage.
However, FCOH gains a magnitude-order reduction of time consumption in updating hash functions, compared to deep-MIHash with either GPU or CPU.
To analyze, deep-MIHash has to accumulate all the available data to re-train the hash functions.
Thus, it avoids the information loss of past streaming data.
However, the training time also linearly grows with the increase of streaming data.
%

\subsection{Limitation and Future Work}
Compared with another inner product based method, BSODH, FCOH is an inner product based scheme with two innovations, \emph{i.e.}, \emph{class-wise updating} and \emph{semi-relaxation optimization}.
It merits in three aspects, \emph{i.e.}, \emph{at least $75\%$ storage saving} (see Sec.\,\ref{class_wise_updating}), \emph{better accuracy and fast online adaptivity} (see Sec.\,\ref{results}) and \emph{training efficiency} (see Sec.\,\ref{training_efficiency}).
Nevertheless, it can only deal with the single-label data due to the design of the class-wise updating scheme, which requires mutually exclusive class labels.
Such a strict requirement might not be satisfied in many cases, \emph{e.g.}, multi-label datasets, which could limit its application in some real-world applications.
More efforts will be made to solve this issue in our future work.

\section{Conclusion} \label{conclusion}
In this paper, we present a novel supervised online hashing method, termed FCOH, which is an inner product based scheme, which addresses the poor online adaptivity and training inefficiency problems that prevail in existing online hashing methods.
To this end, we first develop a class-wise updating scheme that iteratively renews the hash functions at each stage, through which FCOH achieves better performance with much less training data and storage consumption.
Second, we propose a semi-relaxation optimization that relaxes a part of the binary constraints while treating the other part as a constant binary matrix by pre-computing it, through which no extra variables are introduced with less time consumption.
Extensive experiments on three widely-used benchmarks show that FCOH achieves state-of-the-art accuracy and efficiency in the online retrieval tasks.


%




\ifCLASSOPTIONcompsoc
  \section*{Acknowledgments}
\else
  \section*{Acknowledgment}
\fi

This work is supported by the National Natural Science Foundation of China (No. 62025603, No.U1705262, No.61772443, No.61572410, No.61802324 and No.61702136). It is also sponsored by CCF-Baidu Open Fund and Australian Research Council Project FL-170100117.

\ifCLASSOPTIONcaptionsoff
  \newpage
\fi



\bibliographystyle{IEEEtran}
\bibliography{IEEEabrv,main}
%



%

\begin{IEEEbiography}[{\includegraphics[width=1in,height=1.25in,clip,keepaspectratio]{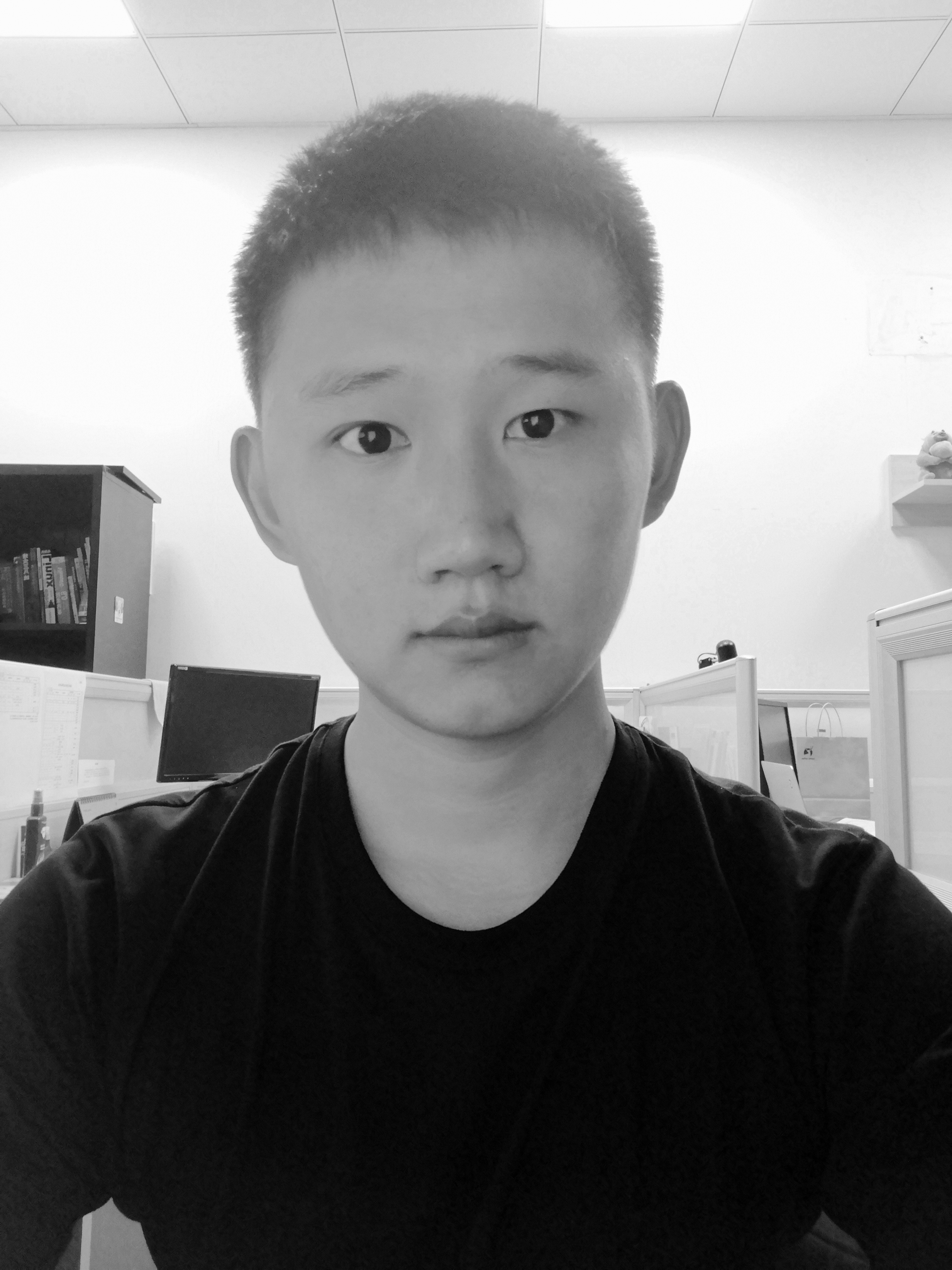}}]{Mingbao Lin}
finished his M.S. study in Xiamen University, China, in 2018. He is currently pursuing the Ph.D degree with Xiamen University, China. His research interests include information retrieval and computer vision.
\end{IEEEbiography}

\begin{IEEEbiography}[{\includegraphics[width=1in,height=1.25in,clip,keepaspectratio]{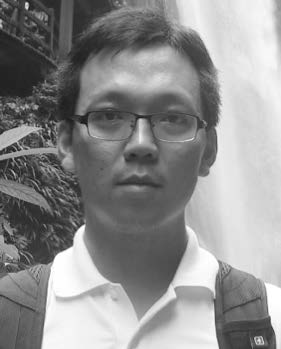}}]{Rongrong Ji}
(SM'14) is currently a Professor and the Director of the Intelligent Multimedia Technology Laboratory, and the Dean Assistant with the School of Information Science and Engineering, Xiamen University, Xiamen, China. His work mainly focuses on innovative technologies for multimedia signal processing, computer vision, and pattern recognition, with over 100 papers published in international journals and conferences. He is a member of the ACM. He was a recipient of the ACM Multimedia Best Paper Award and the Best Thesis Award of Harbin Institute of Technology. He serves as an Associate/Guest Editor for international journals and magazines such as Neurocomputing, Signal Processing, Multimedia Tools and Applications, the IEEE Multimedia Magazine, and the Multimedia Systems. He also serves as program committee member for several Tier-$1$ international conferences.
\end{IEEEbiography}

\begin{IEEEbiography}[{\includegraphics[width=1in,height=1.25in,clip,keepaspectratio]{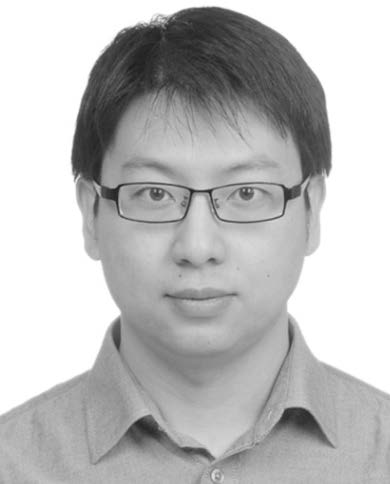}}]{Xiaoshuai Sun}
received the B.S. degree in computer science from Harbin Engineering University, Harbin, China, in 2007, and the M.S and Ph.D. degrees in computer science and technology from the Harbin Institute of Technology, Harbin, in 2009 and 2015, respectively. He was a Post-Doctoral Research Fellow with the University of Queensland from 2015 to 2016 and served as a Lecturer with the Harbin Institute of Technology from 2016 to 2018. He is currently an Associate Professor with Xiamen University, China. He was a recipient of the Microsoft Research Asia Fellowship in 2011.
\end{IEEEbiography}

\begin{IEEEbiography}[{\includegraphics[width=1in,height=1.25in,clip,keepaspectratio]{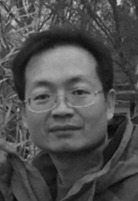}}]{Baochang Zhang}
received the B.S., M.S., and Ph.D. degrees in computer science from the Harbin Institute of the Technology, Harbin, China, in 1999, 2001, and 2006, respectively. From 2006 to 2008, he was a Research Fellow with The Chinese University of Hong Kong, Hong Kong, and also with Griffith University, Brisban, Australia. He is currently a Full Professor with Beihang University, Beijing, China. His current research interests include pattern recognition, machine learning, face recognition, and wavelets.
\end{IEEEbiography}

\begin{IEEEbiography}[{\includegraphics[width=1in,height=1.25in,clip,keepaspectratio]{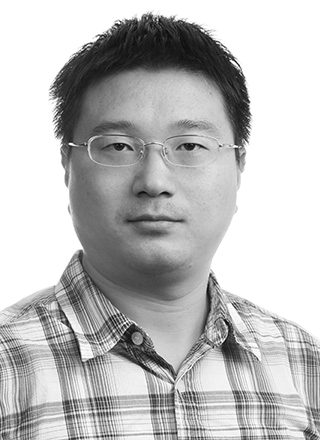}}]{Feiyue Huang}
received the Ph.D degree from Tsinghua University, China, in 2008. He is currently the vice general manager of Tencent Youtu Lab. His research interests include computer vision and pattern recognition.
\end{IEEEbiography}

\begin{IEEEbiography}[{\includegraphics[width=1in,height=1.25in,clip,keepaspectratio]{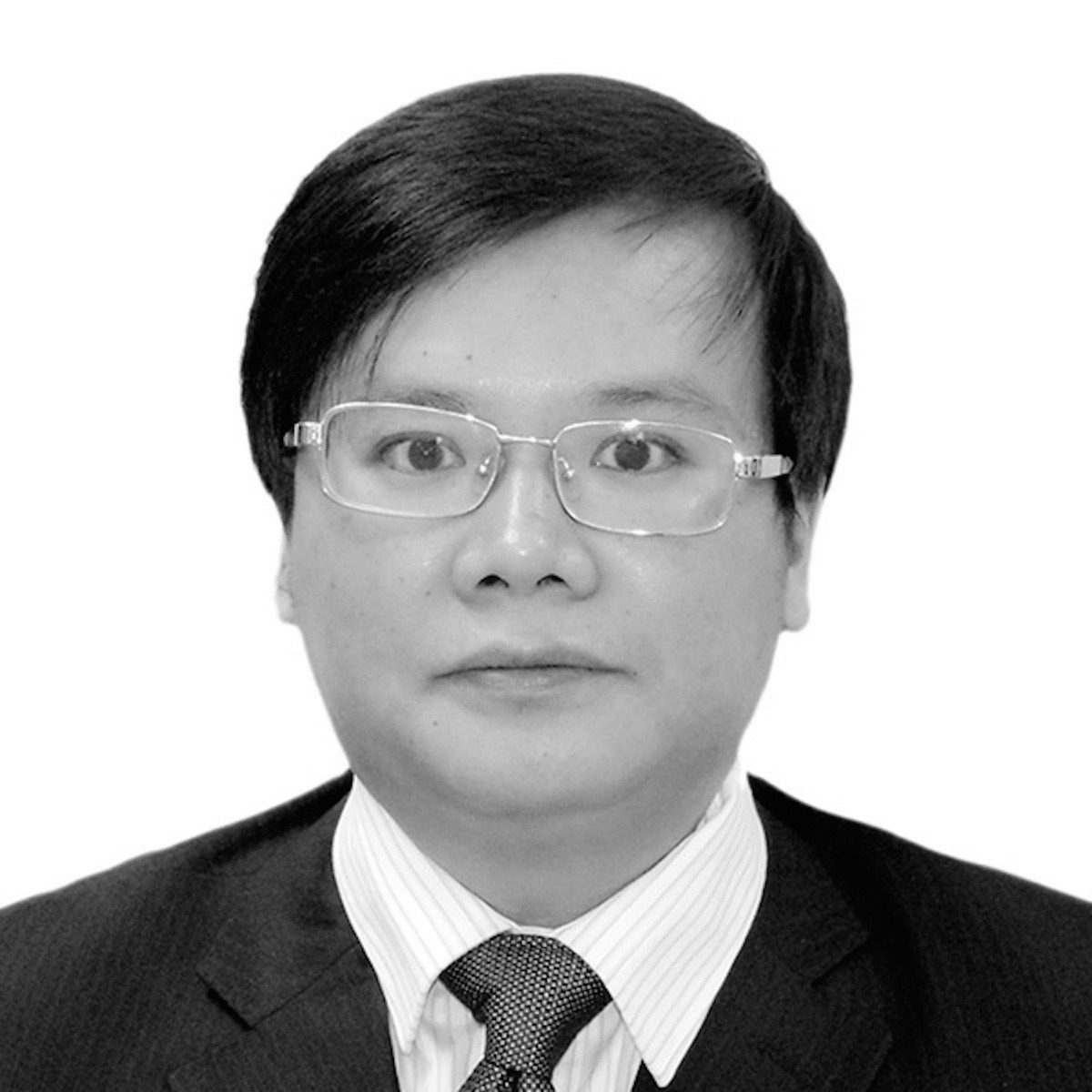}}]{Yonghong Tian}
(S'00-M'06-SM'10) is currently a Boya Distinguished Professor with the Department of Computer Science and Technology, Peking University, China, and is also the deputy director of Artificial Intelligence Research Center, PengCheng Laboratory, Shenzhen, China. His research interests include neuromorphic vision, brain-inspired computation and multimedia big data. He is the author or coauthor of over 200 technical articles in refereed journals such as IEEE TPAMI/TNNLS/TIP/TMM/TCSVT/TKDE/TPDS, ACM CSUR/TOIS/TOMM and conferences such as NeurIPS/CVPR/ICCV/AAAI/ACMMM/WWW. Prof. Tian was/is an Associate Editor of IEEE TCSVT (2018.1-), IEEE TMM (2014.8-2018.8), IEEE Multimedia Mag. (2018.1-), and IEEE Access (2017.1-). He co-initiated IEEE Int’l Conf. on Multimedia Big Data (BigMM) and served as the TPC Co-chair of BigMM 2015, and aslo served as the Technical Program Co-chair of IEEE ICME 2015, IEEE ISM 2015 and IEEE MIPR 2018/2019, and General Co-chair of IEEE MIPR 2020 and ICME2021. He is the steering member of IEEE ICME (2018-) and IEEE BigMM (2015-), and is a TPC Member of more than ten conferences such as CVPR, ICCV, ACM KDD, AAAI, ACM MM and ECCV. He was the recipient of the Chinese National Science Foundation for Distinguished Young Scholars in 2018, two National Science and Technology Awards and three ministerial-level awards in China, and obtained the 2015 EURASIP Best Paper Award for Journal on Image and Video Processing, and the best paper award of IEEE BigMM 2018. He is a senior member of IEEE, CIE and CCF, a member of ACM.
\end{IEEEbiography}

\begin{IEEEbiography}[{\includegraphics[width=1in,height=1.25in,clip,keepaspectratio]{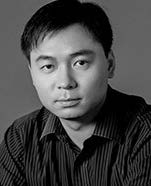}}]{Dacheng Tao} (F'15)
is currently a Professor of Computer Science and an ARC Laureate Fellow in the School of Computer Science and the Faculty of Engineering at The University of Sydney. He mainly applies statistics and mathematics to artificial intelligence and data science. His research is detailed in one monograph and over 200 publications in prestigious journals and proceedings at prominent conferences such as IEEE TPAMI, TIP, TNNLS, IJCV, JMLR, NIPS, ICML, CVPR, ICCV, ECCV, AAAI, IJCAI, ICDM and ACM SIGKDD, with several best paper awards, such as the Best Theory/Algorithm Paper Runner Up Award at IEEE ICDM’07, the Distinguished Paper Award at 2018 IJCAI, the 2014 ICDM 10-year Highest-Impact Paper Award, and the 2017 IEEE Signal Processing Society Best Paper Award. He received the 2015 Australian Scopus-Eureka Prize and the 2018 IEEE ICDM Research Contributions Award. He is a fellow of the Australian Academy of Science, AAAS, ACM and IEEE.
\end{IEEEbiography}




\end{document}